\documentclass[10pt,journal,compsoc]{IEEEtran}

\usepackage{makecell}
\usepackage{hyperref}
\usepackage{array}
\usepackage{graphicx,amssymb,amsmath}
\usepackage{multicol}
\usepackage[noadjust]{cite}
\usepackage{setspace}
\usepackage{subfigure}
\usepackage{graphicx}
\usepackage{float}
\usepackage {url}
\usepackage{stfloats}
\usepackage{amsthm,pifont}
\usepackage{flushend}
\usepackage{cases,subeqnarray}
\usepackage{bm,multirow,bigstrut}
\usepackage{amsmath, amsthm, amssymb}
\usepackage{textcomp}
\usepackage{latexsym,bm}
\usepackage{booktabs}
\usepackage{xcolor}
\usepackage{mathtools}
\usepackage{dsfont}
\usepackage{extarrows}
\usepackage{epsfig}
\usepackage{epsfig}
\usepackage{epstopdf}
\usepackage{algpseudocode}
\usepackage{algorithmicx,algorithm}

\usepackage{helvet}
\theoremstyle{plain}

\theoremstyle{plain}

\usepackage{amsmath}

\IEEEoverridecommandlockouts

\usepackage{fancyhdr}

\begin{document}

\title{Task-Oriented Semantic Communication in Large Multimodal Models-based Vehicle Networks}
\author{Baoxia~Du, Hongyang~Du, Dusit~Niyato,~\IEEEmembership{Fellow,~IEEE}, and Ruidong~Li*,~\IEEEmembership{Senior Member,~IEEE}

\thanks{B.~Du and R. Li (*Corresponding Author) are with the Institute of Science and Engineering, Kanazawa University,
Kanazawa 920-1192, Japan (e-mail: baoxiadu@outlook.com and liruidong@ieee.org)}
\thanks{H. Du is with the Department of Electrical and Electronic Engineering, University of Hong Kong, Pok Fu Lam, Hong Kong (e-mail: duhy@eee.hku.hk).}
\thanks{D. Niyato is with the College of Computing and Data Science, Nanyang Technological University, Singapore (e-mail: dniyato@ntu.edu.sg)}
}

\maketitle

\pagestyle{fancy}
\fancyhf{} 
\fancyhead[C]{\fontsize{7}{9}\selectfont
This article has been accepted for publication in IEEE Transactions on Mobile Computing. This is the author's version which has not been fully edited and \\ content may change prior to final publication. Citation information: DOI 10.1109/TMC.2025.3564543}
\fancyfoot[C]{\fontsize{7}{9}\selectfont © 2025 IEEE. All rights reserved, including rights for text and data mining and training of artificial intelligence and similar technologies. Personal use is permitted,\\ but republication/redistribution requires IEEE permission. See https://www.ieee.org/publications/rights/index.html for more information.}
\fancyhead[R]{\thepage}
\renewcommand{\headrulewidth}{0pt}
\renewcommand{\footrulewidth}{0pt}
\thispagestyle{fancy}

\vspace{-1cm}
\begin{abstract}
Task-oriented semantic communication has emerged as a fundamental approach for enhancing performance in various communication scenarios. While recent advances in Generative Artificial Intelligence (GenAI), such as Large Language Models (LLMs), have been applied to semantic communication designs, the potential of Large Multimodal Models (LMMs) remains largely unexplored. In this paper, we investigate an LMM-based vehicle AI assistant using a Large Language and Vision Assistant (LLaVA) and propose a task-oriented semantic communication framework to facilitate efficient interaction between users and cloud servers. To reduce computational demands and shorten response time, we optimize LLaVA’s image slicing to selectively focus on areas of utmost interest to users. Additionally, we assess the importance of image patches by combining objective and subjective user attention, adjusting energy usage for transmitting semantic information. This strategy optimizes resource utilization, ensuring precise transmission of critical information. We construct a Visual Question Answering (VQA) dataset for traffic scenarios to evaluate effectiveness. Experimental results show that our semantic communication framework significantly increases accuracy in answering questions under the same channel conditions, performing particularly well in environments with poor Signal-to-Noise Ratios (SNR). Accuracy can be improved by 13.4\% at an SNR of 12dB and 33.1\% at 10dB, respectively.

\end{abstract}


\IEEEpeerreviewmaketitle

\section{Introduction}
Large Multimodal Models (LMMs), as one of the significant achievements of Generative Artificial Intelligence (GenAI), are increasingly recognized and utilized in the academic community for their ability to integrate and understand various types of data inputs such as text, images, and videos~\cite{Liu2023OnTH,Lu2023MathVistaEM}. Like humans, LMMs can perceive the real world and make decisions and control tools, playing a crucial role in building universal assistance systems~\cite{Liu2023ImprovedBW,Li2023MultimodalFM,Yan2023GPT4VIW,10571569}. Particularly in AI and Deep Learning (DL), by combining Large Language Models (LLMs) and Computer Vision (CV) technologies, LMMs can comprehensively understand and respond to user needs. For instance, ChatGPT4 and ChatGPT4o~\cite{Achiam2023GPT4TR,openai2024chatgpt}, as forefront products from OpenAI, not only handle textual inputs but also respond to images and voice data, demonstrating impressive interactive and generative capabilities.

Moreover, the applications of LMMs in automated and intelligent systems are expanding~\cite{Cui2023ASO,Zhou2023VisionLM}. In autonomous driving, by integrating data from onboard sensors, real-time traffic information, and driver behavior patterns, LMMs can be utilized to achieve safer navigation and vehicle management. LMMs continuously learn from new data and the interactions with users and environments to adapt to changing driving patterns, user preferences, and evolving road conditions. Through precise fine-tuning or contextual learning to match individual drivers' preferences, LMMs can significantly improve the driving experience~\cite{Cui2023ASO, Ding2023HiLMDTH, Cui2023DriveAY, Jin2023SurrealDriverDG,Zhao2024DriveLLaVAHB}. However, these models face practical challenges such as model scale, data security, and high training and inference costs, which still need addressing to maximize their real-world applications~\cite{Jin2024EfficientML,Liu2024ResourceAI,10588403}.

Deploying LMMs on edge devices such as vehicles presents a significant challenge~\cite{Zhou2023VisionLM}. These models typically have substantial size and demanding computational requirements, with some current LMMs containing from millions to billions of parameters~\cite{Liu2023ImprovedBW,Yin2023ASO}. This necessitates extensive memory and high-speed processors for real-time operation. Due to limited computing resources on edge devices, including processor performance, storage space, and power supply, deploying large-scale models directly on these devices becomes exceedingly difficult~\cite{Shao2024ImpHC}. One strategy to address this issue is to simplify the model’s parameters and reduce complexity by implementing techniques such as model pruning and quantization~\cite{Yuan2023TinyGPTVEM,Zhou2024TinyLLaVAAF,10.1162/tacl_a_00704}, thus creating lighter model versions that require less memory and computational resources, making them more suitable for resource-constrained environments. However, this simplification often sacrifices a degree of model accuracy and performance, which could impact the effectiveness of tasks, especially in applications requiring complex decision-making and deep understanding. Another solution is deploying LMMs on high-performance servers~\cite{Cui2023ASO,10591707,Yang2024PerLLMPI}, which relies on an efficient and stable communication system. 

Benefiting from advancements in the latest communication technologies, such as semantic communication, Sixth Generation (6G) networks, and Vehicle-to-Everything (V2X) networks, deploying complex computational models in vehicles has become feasible. Next-generation communication technologies will offer high data transmission rates and ultra-low latency. These enhancements will significantly improve real-time communication capabilities between vehicles and servers, allowing complex communication computations to be performed instantly~\cite{Wang2023OnTR,Alsharif2020SixthG}. Additionally, progress in semantic communication technology has revolutionized the approach to wireless network communications. Unlike traditional methods that prioritize bit transmission accuracy, semantic communication focuses on transmitting information that is critically important for specific tasks by parsing and understanding the semantic content of data. Recent researches demonstrate the tremendous potential of DL-based semantic communication systems to improve performance and combat noise interference, especially in hostile channel environments~\cite{Zhang2022AUM,Xie2020DeepLE,Huang2023JointTA,Kang2022PersonalizedSI,Du2023AIGeneratedIM}. By developing and optimizing efficient semantic communication models, edge network devices can integrate semantic encoders and decoders, which enhances data processing efficiency and increases the specificity and effectiveness of network communications~\cite{Yang2022SemanticCM,10540315}. 

Regarding the second deployment method, existing works mostly focus on the allocation of LLM computational resources within the vehicle network~\cite{Cui2023ASO,Liu2024ResourceAI}. For example, \cite{Liu2024ResourceAI} investigates the integration of vehicles with LLMs in a 6G environment, where the initial layers of LLM computation are processed on the vehicle side, and the remaining LLM computation tasks are offloaded to Roadside Units (RSUs). By further optimizing resources such as the number of computation layers, transmission power, and GPU frequency, the system's completion time and energy consumption are significantly reduced. 
In~\cite{Chen2024EdgeCloudCM}, researchers propose a flexible framework, EC-Drive, to process problems of varying complexity. For simpler issues such as motion planning, tasks are handled by a lightweight LLM deployed on the vehicle side, whereas more complex problems are uploaded to the cloud to be processed by the more powerful GPT-4. This method effectively reduces inference latency and optimizes the use of communication resources.
However, these works rarely consider the semantic importance of the transmitted data during LMM communication and the impact of noise-induced bit error rates on the accuracy of model responses~\cite{Xie2021TaskOrientedMS,Zhang2022AUM}. When the communication channels between vehicles and servers are poor, a higher error rate will be brought out, which significantly affects the model's performance, potentially leading to erroneous judgments. Therefore, maintaining the efficiency and accuracy of LMMs in complex communication environments is an urgent issue to resolve.

To address these challenges, we propose a task-oriented semantic communication framework for LMMs-based vehicle AI in this paper. Specifically, we employ the Large Language and Vision Assistant (LLaVA) as an LMMs, which innovatively integrates an LLM with a CLIP~\cite{Radford2021LearningTV} visual encoder through a simple projection matrix to construct a visual-language model. This simple architecture has achieved significant success in general visual-language understanding and demonstrated the ability to handle complex question-answering tasks~\cite{Liu2023ImprovedBW,Sun2024SQLLaVASF,Zhao2024DriveLLaVAHB}. We design this framework to optimize data transmission and resource allocation within vehicle AI systems, thereby enhancing the efficiency of data communication and improving the accuracy of the models.

To achieve a more appropriate allocation of transmission resources, we measure the importance of semantic information by integrating both objective and subjective attention. Objective attention highlights key regions in the image, while subjective attention addresses situations where objective regions may not align with the user’s actual focus. By combining these two types of attention, our approach can identify and reliably transmit semantically important image features even under challenging communication conditions, ensuring efficient use of transmission resources and maintaining high response accuracy.

The contributions of this paper are summarized as: 
\begin{itemize}
\item We propose a task-oriented semantic communication framework for vehicle networks, optimized through distributed deployment. A vehicle acts as an encoder that extracts image features and sends them to a cloud server, where computationally intensive tasks are performed. During this process, critical image features are prioritized for transmission. This approach significantly reduces the computational load on the vehicle and enhances the overall security and response speed of the system.
\item We improve the image feature extraction method in the LLaVA model by introducing a task-oriented method to more precisely locate and supplement image features. This improvement significantly reduces the number of image features required for processing, with only a slight sacrifice in model accuracy, thereby speeding up the model's response time.
\item We propose a method that integrates both subjective and objective attention to assess the semantic importance of different patches in an image and to allocate the transmission power for various semantic information accordingly. Furthermore, we optimize the distribution of transmission power among different semantic information. This method significantly improves the model's accuracy while maintaining the same Signal-to-Noise Ratio (SNR).
\end{itemize}

The remainder of this paper is organized as follows: Section~\ref{S2} summarizes related work on LMMs, semantic communication, and visual saliency prediction. Section~\ref{S3} presents the overall design of the task-oriented semantic communication system for vehicle AI. Section~\ref{S4} details the LMM LLaVA and its enhancement method. In Section~\ref{S5}, we introduce methods for predicting objective and subjective attention, as well as a resource allocation method based on the fused attention mechanism for image feature transmission. Section~\ref{S6} analyzes the numerical results. Section~\ref{S7} discusses the key characteristics of the proposed method and outlines future challenges. Finally, Section~\ref{S8} concludes the paper. We summarize the mathematical symbols and explanations in Table~\ref{mathsymbol}.

\begin{table}[ht]
    \centering
    \caption{Summary of Main Symbols.}
    \begin{tabular}{>{\centering\arraybackslash}m{2cm}|p{6cm}}
    \hline
        \textbf{Symbol} & \textbf{Explanation} \\ \hline        
        $X_v$ & The image captured by the vehicle camera. \\  \hline 
        $X_q$ & The user's question. \\  \hline 
        $Z_v$ & Image features. \\  \hline 
        $H_v$ & Visual tokens. \\  \hline 
        $K_q$ & Keyword phrases extracted from the user's question. \\  \hline 
        $g(\cdot)$ & Image encoder model, i.e.,  CLIP-ViT-L.  \\  \hline
        $f_{\phi}(\cdot) $ & LLM backbone used in LLaVA, e.g. Vicuna~\cite{vicuna2023} and Mistral~\cite{Jiang2023Mistral7}. \\  \hline
        $k(\cdot)$ & Word vector model, i.e., GloVe~\cite{Pennington2014GloVeGV}.  \\  \hline
        $P$ & The resolution that the image encoder model can process. \\  \hline

        $\textbf{v}_{K_{q}}$, $\textbf{v}_{c_i}$ & Vector representation of keyword phrases and categories  \\  \hline
        $c_{\text{match}}$, $B_{c_{\text{match}}}$ & The object category that best matches the user's question and its corresponding coordinate box  \\  \hline
        $H_{\text{obj}}$, $H_{\text{sub}}$  & The attention heatmaps of object and subject objects.  \\  \hline
        $\alpha$ & The coefficient for fusing object and subject attention heatmaps.  \\  \hline
        $\beta$ & A weight adjustment variable that adjusts the differences in weight levels between patches.  \\  \hline
    \end{tabular}
    \label{mathsymbol}
\end{table}

\section{Related Work}\label{S2}
In this section, we provide a comprehensive overview of related techniques, including LMMs, semantic communication, and visual saliency. 
\subsection{LMMs}
LLMs like GPT-4~\cite{Achiam2023GPT4TR}, LLaMA~\cite{Touvron2023LLaMAOA}, and Mistral~\cite{Jiang2023Mistral7} have demonstrated exceptional capabilities in understanding textual problems and generating responses. Extending these capabilities, LMMs incorporate visual data into the mix. By integrating visual encoders with LLMs, LMMs can generate chat-like text responses based on images and associated questions, marking significant advancements in visual-language understanding, reasoning, and interactive capabilities~\cite{Xu2024LLaVAUHDAL}. For example, Liu et al.~\cite{Liu2023VisualIT} designed the visual language model LLaVA, which innovatively combines LLMs with the CLIP~\cite{Radford2021LearningTV} visual encoder. Through strategies such as pre-training alignment and targeted instruction adjustments, LLaVA has shown outstanding performance in image-text dialogue tasks. Similarly, Zhu et al.~\cite{Zhu2023MiniGPT4EV} introduced MiniGPT-4, which aligns a frozen visual encoder with the LLM backbone Vicuna~\cite{vicuna2023}, providing advanced visual-language functionalities comparable to those of GPT-4.

Although LLMs have made significant advancements, their large-scale training and deployment incur substantial computational costs. These models typically require extensive data and high-performance hardware resources to achieve high levels of language understanding and generation capabilities. Efforts to address this challenge include Chu et al.~\cite{Chu2023MobileVLMA}, who explored deploying LLMs on mobile devices using architectures and techniques suitable for memory-constrained environments, maintaining high performance. Yuan et al.~\cite{Yuan2023TinyGPTVEM} introduced a smaller LLM, Phi-2, with a unique module for visual and language information fusion, significantly reducing computational costs during model training and inference. Zhou et al.~\cite{Zhou2024TinyLLaVAAF} developed TinyLLaVA, a scaled-down version of the larger LMMs, which explores the effectiveness of various visual encoders, smaller-scale LLMs, and training methods on the performance of these models. TinyLLaVA-3.1B, which combines the Phi-2 LLM and the SigLIP visual encoder, has demonstrated superior performance compared to some existing 7B models. Some research has focused on reducing inference costs through advanced compression techniques like 4/8 bit quantization~\cite{Dettmers2022LLMint88M, Shang2023PBLLMPB}.

Additional studies aim to reduce computational demands by focusing on the efficiency of image data processing. Shang et al.\cite{Shang2024LLaVAPruMergeAT} optimized the selection of visual tokens, dynamically retaining the most critical ones and merging them based on key similarities, achieving an average compression of visual tokens by 14 times compared to the original LLaVA-1.5\cite{Liu2023ImprovedBW}. Ge et al.~\cite{Ge2024ConvLLaVAHB} addressed visual token redundancy by integrating the ConvNeXt~\cite{9879745} visual encoder, which compresses high-resolution images into information-rich visual features, preventing the generation of excessive visual tokens. These studies, however, primarily focus on the visual tokens themselves without much consideration for their relevance to the user's queries. In this paper, we propose a task-oriented image slice localization method that significantly reduces the number of visual tokens the LLM needs to process, thereby saving inference costs while maintaining the model's performance.

\subsection{Semantic Communication}
Semantic communication represents a shift from traditional data transmission methods, focusing on efficiently and securely conveying the essence of messages~\cite{Yang2022SemanticCF,Liang2023GenerativeAS,10571569}. Unlike conventional approaches that emphasize accurate raw data transfer, semantic communication seeks to understand and prioritize the meaning and relevance of the information exchanged. This method aligns more closely with natural human communication, capturing and transmitting the core content rather than duplicating every detail, which enhances efficiency and specificity. A key advantage of semantic communication is its potential to decrease the bandwidth needed for data transfers, reducing transmission costs. By concentrating on meanings pertinent to the final task rather than the exact initial data, semantic systems can achieve higher compression rates~\cite{Xie2020DeepLE}. Furthermore, this approach enhances communication reliability by focusing on the meaning of the information rather than the information itself. The system can maintain message integrity even if parts of the data are corrupted during transmission, ensuring the delivery of crucial information \cite{Xie2023SemanticCW}. The adaptability of semantic communication to assess the importance of different data during transmission is another significant benefit. It allocates transmission power and resources based on the significance of information, prioritizing critical data under suboptimal network conditions~\cite{Yang2022SemanticCF}. This prioritization is particularly vital in scenarios like real-time communication or on devices with limited resources.

Amidst rapid advancements in AI, particularly with the advent of LMMs, researchers are exploring the integration of semantic communication for multimodal data transmission. Studies such as those by Xie et al.~\cite{Xie2021TaskOrientedMS} propose Transformer-based frameworks that optimize data transmission for tasks like image retrieval and visual question answering. These frameworks enhance system robustness, decrease computational complexity, and cut down transmission delays, especially in low SNR environments. Zhang et al.~\cite{Zhang2022AUM} introduced a unified semantic communication system enabled by deep learning, which adjusts the transmission scale based on task demands and channel conditions, significantly reducing overhead and model size while preserving performance. In this paper, we introduce a task-oriented semantic communication framework utilizing the LMM LLaVA. This framework optimizes communication efficiency by understanding and prioritizing the transmission of semantically relevant information to user tasks, ensuring high accuracy and responsiveness in vehicle AI communication scenarios.
\subsection{Visual Saliency}
Visual saliency, a mechanism that allows the human visual system to rapidly detect and recognize critical visual information in complex scenes, plays a crucial role in processing important visual stimuli efficiently. It facilitates the identification of key elements amidst numerous visual inputs, simulating the attention mechanisms inherent in human vision~\cite{Lennie2003TheCO}. Integrating visual saliency with computer vision techniques significantly enhances the efficiency and accuracy of image processing and analysis. For example, in object detection and recognition, saliency detection focuses computational resources on the most informative regions, thereby boosting efficiency and accuracy~\cite{Guo2022SaliencyGD}. During image compression and transmission, prioritizing salient regions ensures the quality of crucial areas, as these contain vital information~\cite{Kang2022PersonalizedSI}. In autonomous driving, saliency detection helps identify critical road elements like pedestrians, traffic signs, and vehicles, enhancing safety~\cite{Hu2020UtilisingVA}.

Recent advances in DL have propelled visual saliency research forward. Kruthiventi et al.~\cite{Kruthiventi2015DeepFixAF} developed a Deep Convolutional Neural Network (DCNN) for saliency prediction, achieving state-of-the-art results by learning hierarchical visual feature representations directly from data. Liu and Han~\cite{Liu2016DHSNetDH} introduced DHSNet, a model that integrates local contextual information and uses a hierarchical recurrent convolutional neural network to progressively refine saliency map details in an end-to-end manner. Pan et al.~\cite{Pan2017SalGANVS} applied a Generative Adversarial Network (GAN) to train a DCNN, involving a generator that creates synthetic samples fitting the data distribution and a discriminator that differentiates between real and synthetic samples. This GAN-based approach significantly improves predictive performance, though it is noted that the generator, based on the VGG-16 architecture, does not consider multi-scale information in images. Tliba et al.~\cite{Tliba2022SATSalAM} proposed Self-Attention Saliency (SAtSal), a novel prediction method that merges high- and low-level features through multilevel skip connections during the decoding phase. SAtSal employs convolutional self-attention modules to integrate multilevel spatial feature signals from the encoder to the decoder network effectively, enhancing the representation of salient regions. Although these visual saliency prediction studies can effectively identify key parts of an image, in real-world scenarios, the user's attention does not always align with the predicted general saliency map~\cite{8444709}. Therefore, we integrate the user's subjective attention into the general saliency map. By combining both types of attention distributions, we can more effectively assess the importance of different regions within an image.

\section{System Model}\label{S3}
In this section, we detail the architecture of the proposed semantic communication system, which is designed to optimize the transmission of visual features in LMMs-based vehicle AI applications. By allocating more resources to transmit higher-priority feature data, the system effectively utilizes bandwidth, enhancing the accuracy with which the AI assistant processes tasks.

To be specific, we explore the implementation of an intelligent driving assistant for vehicles in motion, designed to aid drivers in making quicker, safer decisions based on surrounding environmental information and traffic conditions. The intelligent assistant gathers data through onboard cameras, acquiring detailed information about the vehicle’s surroundings, such as the position of nearby vehicles, the status of traffic lights, and the content of traffic signs, and then synthesizes these observations to offer more meaningful insights. For example, a driver (i.e., user) might inquire: “Which lane should I choose to reach my destination based on the road signs ahead?” or “Is there any obstacle behind me if I reverse now?" The assistant can capture perspectives that are difficult for drivers to carefully observe on their own, and then summarize and present the key information simply and efficiently. This capability facilitates more effective driving and significantly enhances the overall driving experience.

Considering deploying AI, specifically LMMs in vehicles, is costly due to the substantial computational resources and memory required for inference by the underlying LLMs. We adopt an effective solution that offloads these computationally intensive operations to high-performance servers. Consequently, only minimal computing is required on the vehicle side, with only a small amount of data needing to be transmitted from the vehicle to receive responses from the intelligent assistant. This approach alleviates computing workload on onboard systems and enhances processing efficiency. Nevertheless, this method demands an efficient and reliable communication system to support real-time data transmission. Especially in environments with poor quality channels, ensuring the accuracy and security of the transmitted vehicle data becomes critically important. To address these challenges, we introduce a task-oriented semantic communication framework based on the LMM LLaVA model. 
Considering the high computational demands of LMMs, where the LLM part typically represents the most computationally expensive component~\cite{Shang2024LLaVAPruMergeAT}, the scale of visual encoders is generally smaller in comparison. For instance, the widely used CLIP visual encoder ViT-L has about 300 million parameters, whereas corresponding LLMs such as LLaMA can have up to 700 million or even 1.3 billion parameters~\cite{Touvron2023LLaMAOA}. Therefore, we opt to deploy the smaller-scale visual encoder module on the vehicle side.
The system only needs to transmit encoded image features, as opposed to directly transmitting the complete image data captured by the vehicle camera. This can effectively reduce the risk of leaking sensitive information (e.g., in a vehicle network environment, the image may contain personal information such as the faces of the vehicle owner or pedestrians. Transmitting processed feature data can prevent such information from being directly exposed), thereby protecting user privacy and the security of the system's data. During the data transmission process, the system evaluates the importance of each semantic feature by merging objective environmental attention with the user’s subjective attention, and dynamically allocates transmission power based on semantic importance. This means that more transmission resources are allocated to features identified as more important or critical, ensuring their integrity and priority during transmission. For instance, for an important traffic sign, the system increases the transmission power for that data segment to reduce the likelihood of data loss and ensure accurate transmission even in poor communication environments.

\begin{figure*}[ht]
  \centering
  \includegraphics[width=0.9\textwidth]{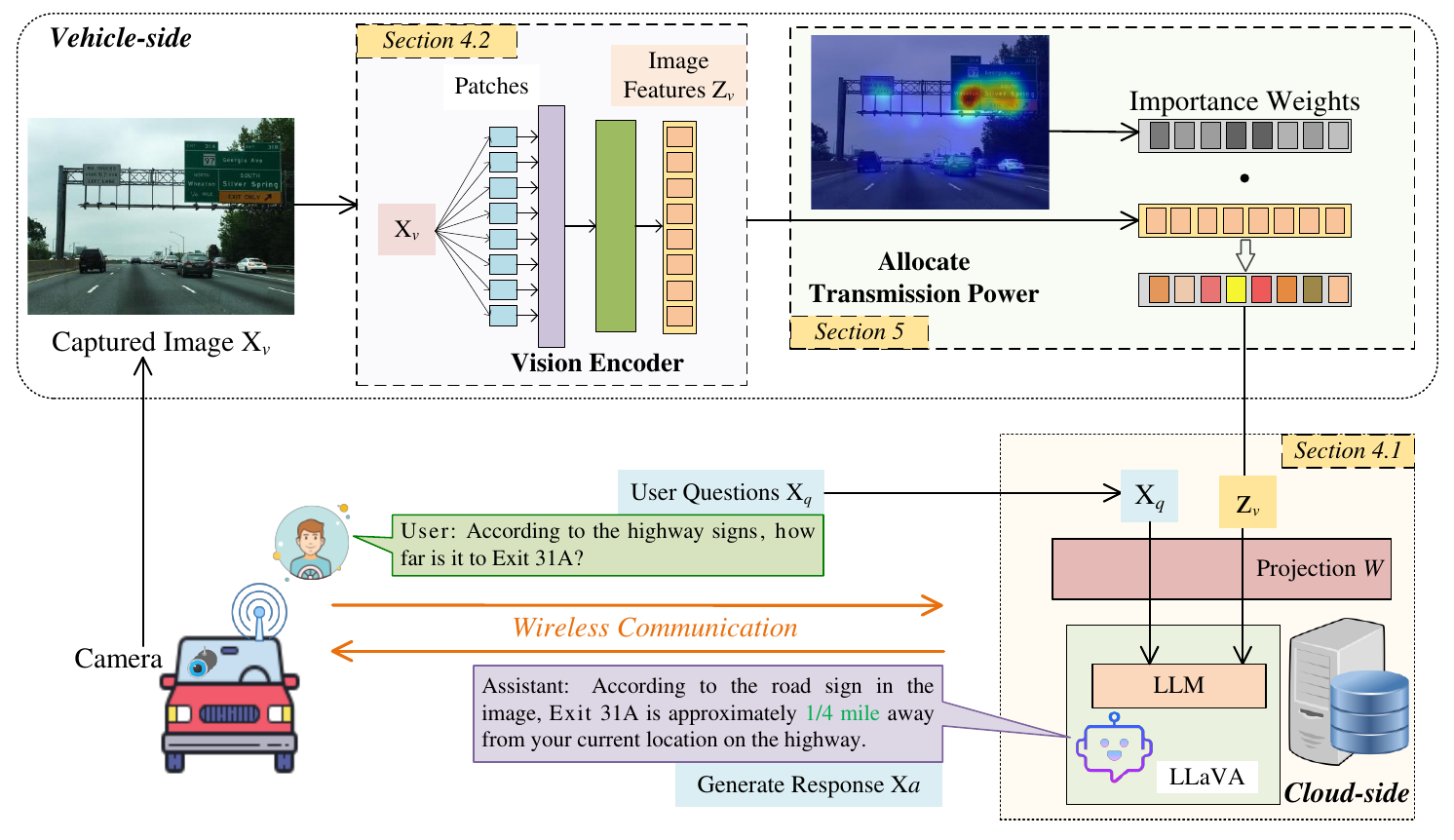}
  \caption{The proposed task-oriented semantic communication system framework. Section~\ref{S41} introduces the architecture of LLaVA, Section~\ref{S42} presents the optimization method for image encoding, and Section~\ref{S5} details the proposed fusion attention-based power allocation method.}
  \label{systemmodel}
\end{figure*}

\begin{figure}[htbp]
  \centering
  \includegraphics[width=0.45\textwidth]{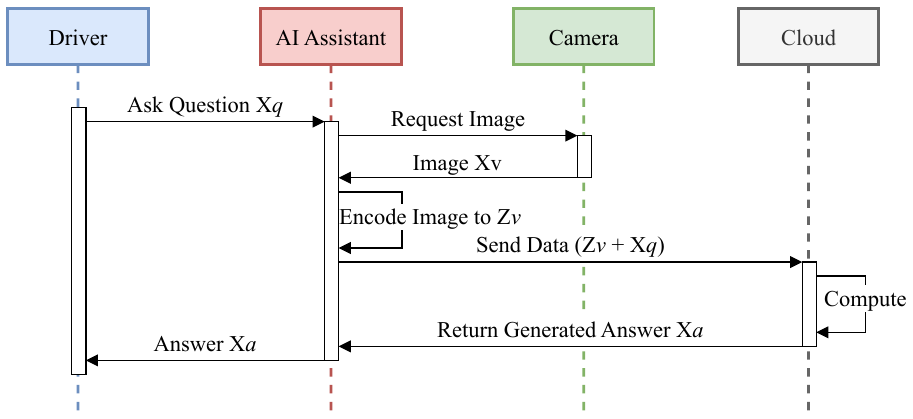}
  \caption{A simple processing flow of the system framework}
  \label{flow}
\end{figure}

The system model architecture is shown in Fig.~\ref{systemmodel}. The process begins with capturing an image \(X_v\) from a camera installed on the vehicle, which serves as input to the visual encoder. The visual encoder's task is to extract relevant image features $Z_v = g(X_v)$,  which are essentially high-level representations of the image. The extracted image features $Z_v$ are then combined with the user's textual information \(X_q\) and transmitted to the cloud server through a wireless communication network. During transmission, the system generates an attention heatmap by merging objective attention and the user's subjective attention toward the image, assigning importance weights to each feature of \(Z_v\). Parts with higher importance are allocated more power to ensure accurate transmission of crucial information. Subsequently, the server receives the semantic information and processes it further, transforming \(Z_v\) through a trainable projection matrix $W$ into language embedding tokens $H_v$, aligning them correctly with the word embedding space to feed into LLM. The LLM then generates a response \(X_a\) to answer the user's question. Fig.~\ref{flow} provides a simplified sequence diagram of this process, illustrating each step from initial data capture to final answer delivery. Although our framework is demonstrated using the LLaVA model, its principles of semantic-driven transmission and task-oriented resource allocation can be readily applied to a broad range of similar LMM architectures that adopt patch-based image encoding, including widely used CLIP-ViT~\cite{Radford2021LearningTV} and SigLIP~\cite{zhai2023sigmoid}.

\section{The Optimized Model LLaVA-SM}\label{S4}
In this section, we first introduce the LLaVA model and its method for slicing high-resolution images. Following that, a task-oriented method for image slicing is proposed to address the problem of excessive visual tokens, which aims to optimize the processing of visual data by ensuring that only the most semantically relevant segments of an image are retained for further analysis.
\subsection{The Architecture and Image Slicing Methods in LLaVA}\label{S41}
\textbf{The Architecture of the LLaVA. }
LLaVA represents a significant advancement in open-source LMMs, exemplifying a novel approach to end-to-end training for general visual and language understanding, achieving impressive chat capabilities. LLaVA employs a simple projection matrix to link a pre-trained visual encoder with an LLM, enabling the LLM to comprehend images and answer questions about their content. The chosen LLM backbone for LLaVA, the Vicuna~\cite{vicuna2023} model  \( f_{\phi}(\cdot) \), is distinguished among publicly available models for its superior capability to follow language-based instructions. When processing an input image \( X_v \), LLaVA utilizes the pre-trained visual encoder CLIP-ViT-L-336px  (the highest resolution available for CLIP) to extract visual features \( Z_v = g(X_v) \). To connect these image features to the word embedding space, LLaVA applies a trainable linear layer, specifically the projection matrix  \( W \), which transforms the visual features \( Z_v \) into language embedding tokens $H_v$ that match the dimensions of the word embeddings used in the language model. This results in a sequence of visual tokens $H_v$. This process can be represented as:
\begin{equation}
    H_v = W \cdot Z_v, \text{ with } Z_v = g(X_v).
\end{equation}

LLaVA adopts one of the simplest architectural designs within LMMs, requiring only a simple fully connected projection layer to be trained on approximately 600,000 image-text pair data. The process of LLaVA effectively demonstrates how visual inputs are transformed into visual features that an LLM can understand and respond to. This model handles multimodal data and achieves precise performance on visual instruction following tasks. The method of integrating visual and language information positions LLaVA as a potent tool in the multimodal AI field, effectively handling complex tasks that require the fusion of visual and linguistic information~\cite{Shi2024MathLLaVABM}.

\textbf{Image Slicing Methods in LLaVA. }Visual encoders serve as the foundation for LMMs to understand the visual world, mapping image data into visual features comprehensible by LLMs. Most existing LMMs perceive images at fixed resolutions, such as InstructBLIP at $224 \times 224$ and LLaVA-1.5 at $336\times 336$. This basic compression method, which disregards the original image size, often results in significant shape distortion and blurring of the image content. Such issues severely impair the LMMs' ability to perform fine-grained image recognition, such as in tasks involving Optical Character Recognition (OCR) and understanding of small objects, potentially exacerbating model hallucinations and incorrect responses.

\begin{figure*}[ht]
  \centering
  \includegraphics[width=0.98\textwidth]{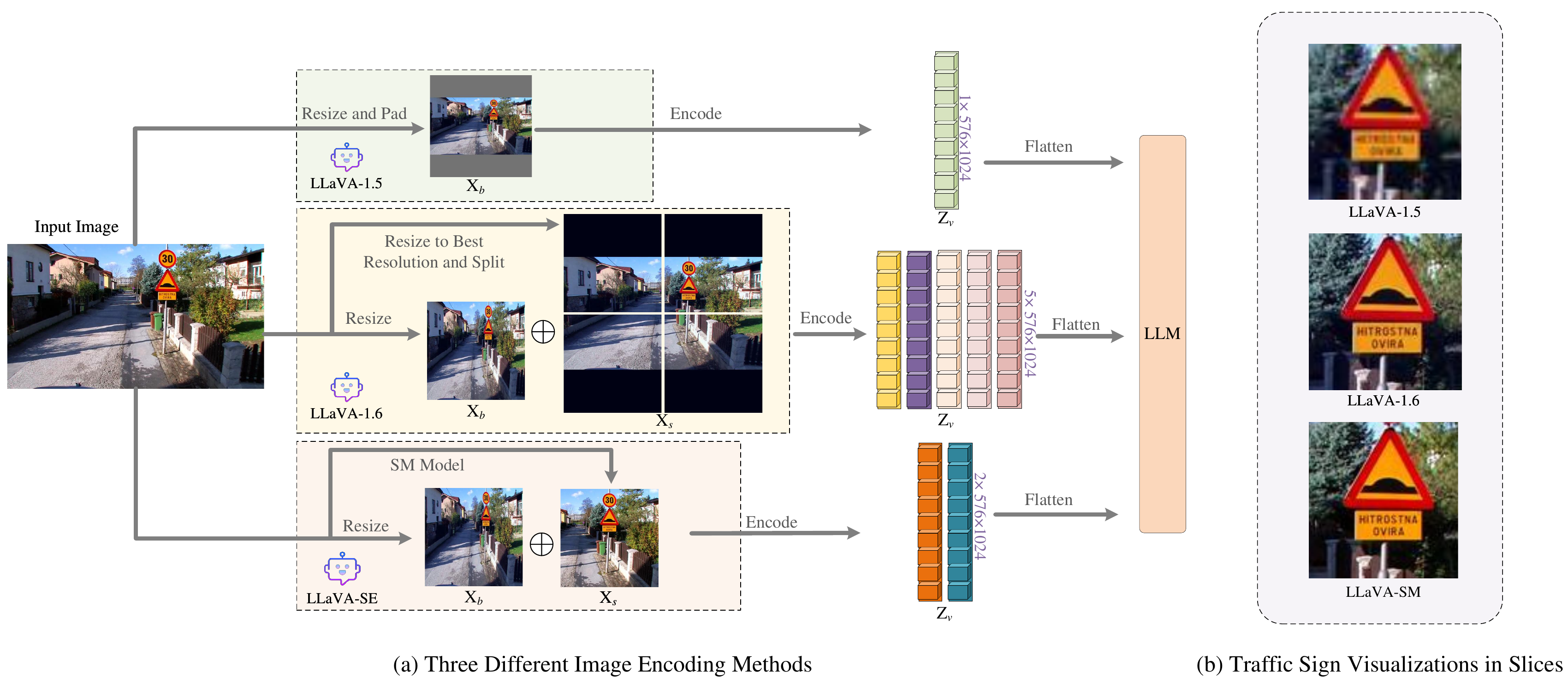}
  \caption{Three different image encoding mechanisms (i.e., LLaVA-1.5, LLaVA-1.6, and the proposed LLaVA-SM). The visualization displays the clarity achieved by each method in critical areas of the image.}
  \label{feature selected}
\end{figure*}

As shown in Fig.~\ref{feature selected}, LLaVA-1.5 uses the pre-trained CLIP-ViT-L/14 model with an image input size fixed at $336 \times 336$ pixels. This model adopts a method of proportionally compressing the image and padding its shorter side to adjust the size to $336 \times 336$. However, this processing can lead to blurred or even lost image details during compression, particularly affecting performance of the OCR tasks, where the model may fail to deliver correct responses. To address this issue, in LLaVA-1.6, the authors introduce a more flexible approach to handling high-resolution images, aimed at enhancing the model's capability to recognize image detail features. This method is not limited to merely compressing the original image but involves dividing the image into a base image $X_b$ and image slices $X_s$. The base image $X_b$ is processed similarly to LLaVA-1.5, directly downsampling the original image to $336 \times 336$ pixels. For image slices, the model first scales the original image based on its resolution and pre-set grids to a maximum resolution of 672×448 (or 448×672), then determines the number of slices per direction based on the scaled image’s resolution and aspect ratio. Each slice $X_s$ is then padded to the target resolution of $336 \times 336$. Notably, before being fed into the LLM, features from the paddings are discarded to improve efficiency. This innovative improvement significantly enhances LLaVA-1.6’s performance in complex visual tasks, especially in applications requiring high detail recognition. While this method considerably improves model accuracy, the addition of image slices leads to a substantial increase in the number of visual tokens that the LLM must process (e.g., a 672 × 1008 resolution image will produce 3,456 visual tokens for LLaVA-1.5~\cite{Liu2023ImprovedBW, Xu2024LLaVAUHDAL}), greatly increasing the model’s inference cost and response time.

\subsection{The Proposed Image Slicing Method}\label{S42}
To address the issue of excessive visual tokens generation caused by image slicing, we propose a Semantic Matching (SM)-based image slicing selection method, which involves three basic lightweight models, i.e., YAKE~\cite{Campos2018ATF}, You Only Look Once (YOLO)~\cite{ultralytics2023yolo} and GloVe~\cite{Pennington2014GloVeGV}. In real-world road traffic scenarios, where user queries often concern elements such as traffic signs, vehicles, and pedestrians, not all parts of an image are equally important. Therefore, as shown in Fig.~\ref{feature selected}, our approach introduces a lightweight SM technique that identifies key areas within an image, allowing the system to encode only the details of regions of interest to the user. This significantly reduces the number of image slices and, consequently, the number of image tokens the LLM needs to process, thus enhancing response speed. This process is primarily divided into three steps: Keyword Extraction, Object Detection, and Semantic Matching. The steps are described in detail below.

\textbf{Keyword Extraction: }The system first needs to obtain the keywords from the user's question $X_q$, which is achieved through the YAKE~\cite{Campos2018ATF} algorithm. YAKE is a classic unsupervised keyword extraction algorithm capable of efficiently identifying important keywords from natural language queries. The YAKE algorithm comprises four principal components. Initially, the text preprocessing stage involves tokenizing the input text using punctuation and special characters. Subsequently, in the feature extraction stage, word weights $\mathrm{S}({w})$ are assigned to each word $w$ based on five specific attributes: Casing $\mathrm{W}_{\text {Case}}$, Word Position $\mathrm{W}_{\text {Position}}$, Word Frequency $\mathrm{W}_{\text {Freq}}$, Word Relatedness to Context $\mathrm{W}_{\text {Rel}}$, and Word Differentiation by Sentence $\mathrm{W}_{\text {DifSentence}}$. Lastly, the calculation of $\mathrm{S}({w})$ integrates these five evaluation metrics to determine each word's significance as follows:

\begin{equation}
    \mathrm{S}({w})=\frac{\mathrm{W}_{\text {Rel }} \cdot \mathrm{~W}_{\text {Position }}}{\mathrm{W}_{\text {Case }}+\frac{\mathrm{W}_{\text {Freq }}}{\mathrm{W}_{\text {Rel }}}+\frac{\mathrm{W}_{\text {DifSentence }}}{\mathrm{W}_{\text {Rel }}}}.
\end{equation}
In this model, words with lower scores are deemed more important. During the candidate keyword generation stage, sequences up to three words in length are generated using a sliding window technique, excluding sequences that begin or end with stop words. The final score of the keywords $\mathrm{S}(Kw)$ is based on the weight of individual terms $\mathrm{S}({w})$ and their total frequency $\operatorname{TF}(Kw)$, computed as follows:
\begin{equation}
    \mathrm{S}(Kw)=\frac{\prod_{w \in Kw} S(w)}{\operatorname{TF}(Kw) \cdot\left(1+\sum_{w \in K w} S(w)\right)}.
\end{equation}
where $Kw$ represents a candidate keyword with a maximum scale of 3 terms. Thus, a lower score $\mathrm{S}(Kw)$ for each candidate keyword $Kw$ correlates with higher importance. Through these steps, the YAKE algorithm effectively extracts the most critical keyword phrases $K_{q}$ and its corresponding scores $S_q$, represented as $Y = \{(K_{q}, S_q)\}$, from the text~\cite{Campos2018ATF}. This process can be represented as: $K_q \leftarrow {\text{YAKE}}(X_q).$ 

\textbf{Object Detection: }Following the initial keyword extraction with YAKE, the system employs a pre-trained YOLO~\cite{ultralytics2023yolo} object detection model to identify objects within the image. The YOLO model categorizes objects detected in the image and labels them with bounding boxes, producing a set of detection results $O = \{(c_i, B_i)\}$, where $c_i$ represents the category label and $B_i$ is the bounding box.

\textbf{Semantic Matching: }The system then matches the keyword phrase $K_{q}$ extracted by YAKE with the object labels detected by YOLO. This process is accomplished by computing the semantic similarity between the keywords and the object labels using cosine similarity. 
Specifically, given a keyword phrase $K_{q}$, it consists of several words $w_1$,$w_2$, ..., $w_n$. For each word $w_i$, its vector representation $\textbf{v}_{w_i}$ is obtained from a pre-trained lightweight word vector model GloVe~\cite{Pennington2014GloVeGV} $k(\cdot)$, which can be represented as $\textbf{v}_{w_i} = k(w_i)$, and the vector representation $\textbf{v}_{K_{q}}$ of $K_{q}$ is the average of the vectors of the words it contains. $\textbf{v}_{K_{q}}$ can be obtained by
\begin{equation}\label{math4}
    \textbf{v}_{K_{q}} = \frac{1}{n} \sum_{i=1}^{n} \textbf{v}_{w_i},
\end{equation}
where $n$ is the number of words contained in the keyword phrase $K_{q}$. The similarity $\text{Sim}(\mathbf{v}_{K_q}, \mathbf{v}_{c_i})$ between the two phrases is obtained through cosine similarity, calculated as follows:

\begin{equation}\label{math5}
    \text{Sim}(\mathbf{v}_{K_{q}}, \mathbf{v}_{c_i}) = \frac{\mathbf{v}_{K_{q}} \cdot \mathbf{v}_{c_i}}{\|\mathbf{v}_{K_{q}}\| \|\mathbf{v}_{c_i}\|},
\end{equation}
where $\mathbf{v}_{K_{q}}$ is the vector representation of the keyword phrase $K_{q}$, $\mathbf{v}_{c_i}$ is the vector of the $i$-th category in the detection results $O$. and $\| \cdot\|$ represents the norm of the vector. Subsequently, the system finds the highest similarity match among multiple keywords $K_{q}$ and category label $c_i$ by:

\begin{equation}
    (c_{\text{max}}, k_{\text{max}}) = \underset{K_{q} \in K, c_i \in C}{\text{argmax}}  \text{Sim}(K_{q}, c_i),
\end{equation}
where $K$ is the set of keyword phrases extracted from $X_q$ and $C$ is the set of category labels. The system then identifies the category with the highest similarity to $K_q$ and retrieves the corresponding bounding box positions as: 
\begin{equation}\label{match1}
    c_{\text{match}} = c_{\text{max}},
\end{equation}
\begin{equation}\label{match2}
B_{c_{\text{match}}} = \{B_i \mid (c_i, B_i) \in O, c_i = c_{\text{match}}\}.
\end{equation}

Through this method, the system identifies the objects of user interest—those objects in the image that correspond most closely to the extracted keywords. By matching all detected categories $c_i$,  the position boxes of these targets $B_{c_{\text{match}}}$ within the image are determined. The whole process can be represented as:
\begin{equation}\label{SM1}
    B_{c_{\text{match}}} \gets SM(X_v, O).
\end{equation}
Furthermore, a flexible target area cropping method $\mathcal{F}_s(X_v,B_{c_{\text{match}}}, P)$ is defined, which selects appropriate target area for image slice based on the position and number of target boxes. The slice is then rescaled to the resolution $P \times P$, enabling it to be processed by the image encoder. The overall algorithm of the SM module is presented in Algorithm \ref{SM}. 

\begin{algorithm}[htb]
\caption{Semantic Matching (SM) Model}\label{SM}
\begin{algorithmic}[1]

\newcommand{\Input}[1]{\State \textbf{Input:} #1}
\newcommand{\Output}[1]{\State \textbf{Output:} #1}

\Input{Captured image $X_v$ on camera, user question $X_q$, patch size $P$}
\Output{Image slice $X_s$}

\Procedure{Feature-Selected}{$X_v, X_q$}
    \State Extract keywords $K_q$ from $X_q$ using YAKE: 
    \begin{center}
        $K_q \gets \text{YAKE}(X_q)$
    \end{center}
    \State Detect objects in $X_v$ using YOLO: 
    \begin{center}
        $O = \{(c_i, B_i)\} \gets \text{YOLO}(X_v)$
    \end{center}

    \State Calculate the cosine similarity between $K_q$ and $c_i$ according to~\eqref{math4} and~\eqref{math5}
    \State Obtain object boxes $B_{c_{\text{match}}}$ corresponding to $K_q$ using~\eqref{match1} and~\eqref{match2} 

    \State $X_s \gets \mathcal{F}_s({X_v, B_{c_{\text{match}}}, P})$ 
    \State \Return $X_s$
\EndProcedure

\end{algorithmic}
\end{algorithm}

\section{Fusion attention-based power allocation scheme}\label{S5}

As discussed previously, when users interact with an intelligent assistant using images, their queries often focus on specific critical elements within the image. For instance, when a user asks about the status of a traffic light ahead, the visual information about the traffic light represents only a small part of the entire image. Therefore, ensuring the accurate transmission of this particular area is essential. Precisely transmitting these key visual features and reducing the error rate of image semantic features will directly impact the correctness of the model's response. However, with poor channel quality or limited bandwidth, these key visual features may compete for resources with other features to be transmitted, affecting the accuracy of the transmission. To address this challenge, we propose a Fusion Attention-based Semantic Communication (FA-SemCom) framework, which quantifies the importance of features in different image patches by merging user objective and subjective attention. This method further enables dynamic resource allocation to ensure the quality of critical semantic information transmission. This section introduces the methods for predicting both objective and subjective user attention and the implementation of dynamic transmission power. 

\subsection{Objective Attention}
In this study, the objective visual attention prediction function $f_{\text{obj}}(\cdot)$ utilizes a pre-trained lightweight convolutional encoder-decoder network model~\cite{8444709}, specifically designed to capture salient features in images. This allows for the effective extraction of key information from any given visual input $X_v$. As a result, the saliency map, i.e., the objective visual attention heatmap $H_{\text{obj}}$, can be represented as:
\begin{equation}\label{obj}
H_{\text{obj}}= f_{\text{obj}}(X_v).
\end{equation}

The saliency prediction model is structured into three main components: Encoder, Atrous Spatial Pyramid Pooling (ASPP), and Decoder~\cite{8444709}. The image encoder uses a modified VGG16 backbone with pooling layers removed to extract deep features from the image. The ASPP module is employed to capture features at different spatial scales concurrently, enhancing the model's ability to discern relevant spatial details. Finally, the decoder network reconstructs these features to restore the input image size, resulting in the generation of the saliency map. This architecture precisely identifies visually important regions and maintains high efficiency and accuracy when processing complex images. The lightweight design of the network ensures fast processing speeds, making it suitable for real-time application scenarios. 

\subsection{Subjective Attention}
Most visual saliency models are primarily focused on predicting generic saliency maps, which, in certain scenarios, may not accurately reflect the image regions corresponding to the user's specific concerns~\cite{8444709}. In other words, the areas of interest to the user do not always align with the locations of highest visual saliency. To address this issue, we propose a method for predicting user subjective attention, enabling the system to accurately identify image regions relevant to the user’s query.

The implementation of this subjective attention is facilitated through the SM module proposed in Section ~\ref{S4}, which precisely determines the detailed bounding box coordinates $B_{c_{\text{match}}}$ of the areas of interest to the user. These regions are then assigned higher attention weights, and a subjective attention heatmap is subsequently generated. This process can be mathematically represented as:
\begin{equation}\label{sub}
    H_{\text{sub}} = f_{\text{sub}}(X_v, B_{c_{\text{match}}}),
\end{equation}
where~$B_{c_{\text{match}}}$ can be obtained by~\eqref{match2}.

This connection between user inquiries and visual components ensures that the attention mechanism focuses on specific areas of the image that directly respond to the user's questions, rather than merely highlighting generally salient features.

\subsection{The Design of FA-SemCom}
After obtaining the user's objective attention heatmap $H_{\text{obj}}$ and subjective attention heatmap $H_{\text{sub}}$, we introduce an attention fusion coefficient $\alpha$ to balance the weights of objective and subjective visual attentions. The fusion process can be represented by

\begin{equation}
H_a = \alpha \cdot H_{\text{obj}} + (1 - \alpha) \cdot H_{\text{sub}},
\end{equation}
where $H_a$ is the combined user attention heatmap, and $\alpha$ is a value within the range $[0, 1]$. In this study, we set $\alpha$ to 0.5, aiming for an equitable blend of both attention types. 

\begin{figure*}[t]
  \centering
  \includegraphics[width=0.9\textwidth]{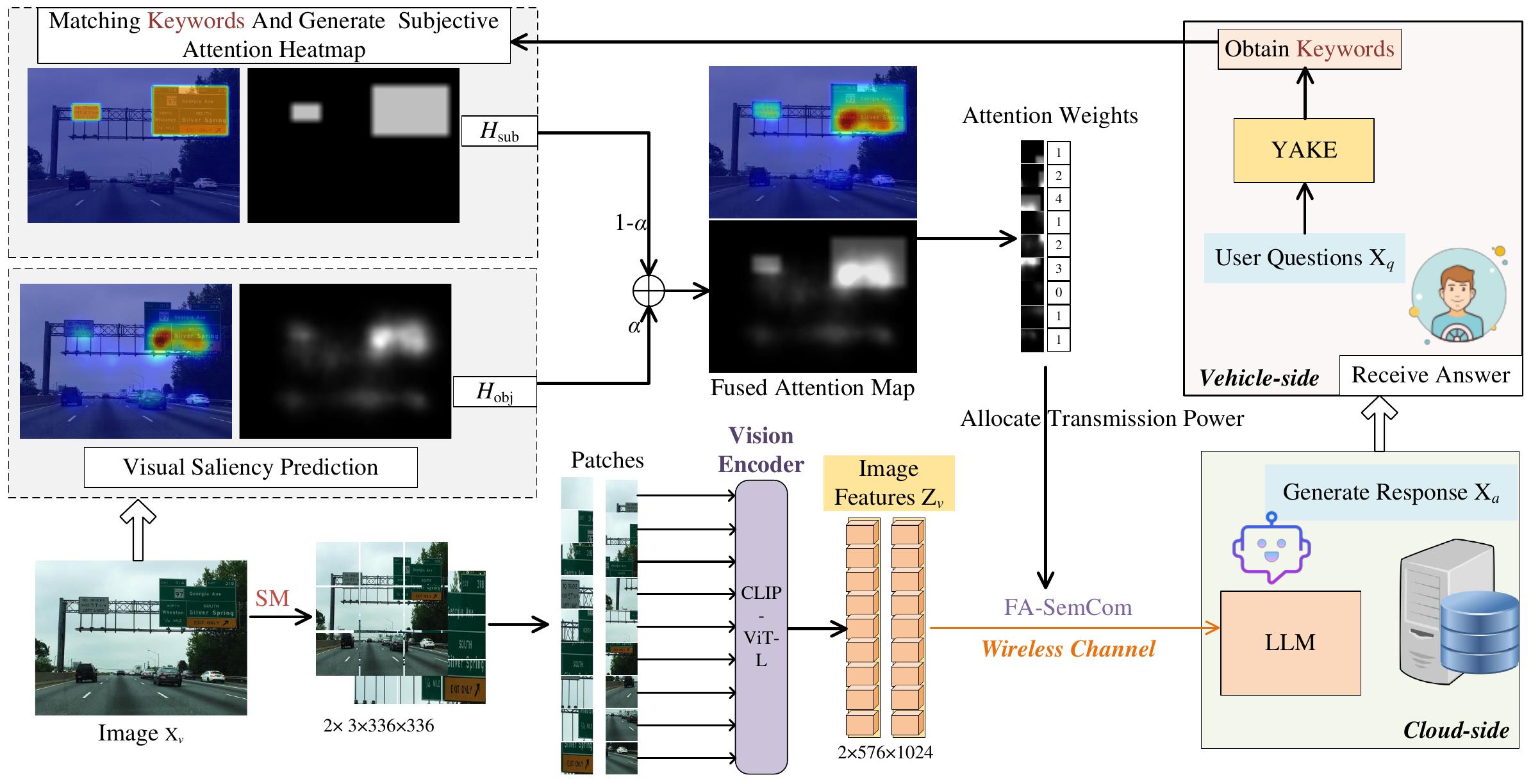}
  \caption{The proposed fusion attention-based resource allocation method. By evaluating the attention weights of each image patch, resources are concentrated on the critical parts of the image.}
  \label{attentionmap}
\end{figure*}

Next, the question is how to use $H_a$ to allocate weights to the encoded image features. To address this issue, we first introduce the image encoding method of LLaVA. In the LLaVA model, the image encoder is a Vision Transformer (ViT), and the encoding process is illustrated in Fig.~\ref{attentionmap}. In particular, given an input image ${X} \in \mathbb{R}^{H \times W \times C}$, where $(H, W)$ represents the original image resolution and $C$ denotes the number of channels in the image, the preprocessing step involves dividing ${X}$ into multiple blocks, each of size $S$. These blocks, referred to as patches, are then flattened into a 2D sequence, represented as ${X}_p \in \mathbb{R}^{N \times (S^2 \cdot C)}$, with the total number of patches derived from the original image being $N=HW/S^2$. These patches are projected through a linear layer to map the dimensions from $(S^2 \cdot C)$ to $D$, maintaining the number of image patches $N$ unchanged. The mapped patches, ${X}_p \in \mathbb{R}^{N \times D}$, are then fed into a multi-layer Transformer encoder. The Transformer processes these image embeddings using self-attention mechanisms and multi-layer perceptron blocks. Ultimately, the ViT-encoded image features $Z_v$ are output.

Motivated by the principle of image encoding by ViT into patches, we propose a method based on fused visual attention to allocate transmission power differentially across patch features. Using the previously obtained user-centric fused attention map, different visual importance weights $W_p$ can be assigned to each of the $N$ patches in the image. This method of fusing visual attention allows the system to specifically identify the most critical patches of the image and subsequently develop an appropriate transmission power allocation strategy. This ensures an efficient use of communication resources and optimizes the transmission efficiency, guaranteeing the accurate transfer of crucial information. Each patch can be denoted as \( p_i \), where \( i = 1, 2, \ldots, N \), the visual importance weight for each patch can be expressed as:
\begin{equation}
    W_p(p_i) = H_a(p_i).
\end{equation}

To simplify the calculation process, the importance weights $W_p(p_i)$ of patches are quantized into $L$ distinct levels. Additionally, we set the importance level $W_L$ of the patches with $W_p = 0$ to 1 to highlight the distinctions from other patches with higher importance. The quantization process is defined as follows:

\begin{equation}\label{rank}
W_L(p_i) =
\begin{cases} 
1 & \text{if } W_p = 0 \\
\left\lceil \text{Norm}(W_p(p_i)) \cdot (L-1) \right\rceil + 1 & \text{if } W_p > 0 
\end{cases},
\end{equation}
where $\text{Norm}(W_p(p_i))$ refers to a normalization function that scales the weights $W_p(p_i)$ to a $[0,1]$ range. Motivated by the design of a user-object-attention level dataset~\cite{9999298}, we set $L=5$ in this study. Following the quantization of each patch's weight \( W_L(p_i) \), the transmission power is allocated based on these quantized weights:
\begin{equation}
P(p_i) = \frac{(W_L(p_i))^\beta}{\sum_{j=1}^{N} (W_L(p_j))^\beta} \times P_{\text{total}},
\end{equation}
where $\beta$ is a tunable parameter that adjusts the sensitivity of the power allocation to differences in weight levels between patches. A higher $\beta$ value amplifies the differences in power allocation corresponding to the weight levels, making the disparity more pronounced; conversely, a lower $\beta$ value smooths out these differences.
Ultimately, patches feature with higher importance weights are allocated more transmission power, ensuring they retain higher fidelity during transmission. The overall process of resource allocation is presented as Algorithm \ref{Resource Allocation}. This method effectively maximizes the impact of data transmission by focusing resources on the most critical parts of the image and ensures the accurate transfer of important data by reserving bandwidth for the most significant patches.

\begin{algorithm}
\caption{Resource Allocation in FA-SemCom}\label{Resource Allocation}
\begin{algorithmic}[1]

\newcommand{\Input}[1]{\State \textbf{Input:} #1}
\newcommand{\Output}[1]{\State \textbf{Output:} #1}

\Input{Captured image $X_v$, detection results $O$, patch size $P$}
\Output{Server receives image features $Z_v$}

\Procedure{Resource Allocation}{$X_v, X_q$}
    \State Obtain the objective attention heatmap of $X_v$ according to~\eqref{obj}
    \State Get the box coordinates from the SM module: 
    \begin{center}
        $B_{c_{\text{match}}} \gets SM(X_v, O)$
    \end{center}
    \State Generate the subjective attention heatmap of $X_v$ according to~\eqref{sub}
    \State Generate the fused attention map $H_a$:
    \begin{center}
        $H_a = \alpha \cdot H_{\text{obj}} + (1 - \alpha) \cdot H_{\text{sub}}$
    \end{center}
    \State Obtain the importance weight $W_p(p_i)$ for each patch and quantize it using~\eqref{rank}
    \State $X_b \gets \text{Resized}(X_v, P^2)$
    \State $X_s \gets SM(X_v, O)$, then resized to $P^2$
    \State $X_b$ and $X_s$ are separately converted to feature vectors $Z_v$ of length $N$ by the image encoder
    \State Allocate transmission power to each dimension of the image feature according to $W_p(p_i)$~\cite{Kang2022PersonalizedSI}
    \State The vehicle sends the image features $Z_v$ to the server
\EndProcedure

\end{algorithmic}
\end{algorithm}

\section{Numerical Results}\label{S6}

\subsection{Dataset and Environment}

The experimental setup was configured on a standard Ubuntu 20.04 system, equipped with a NVIDIA RTX A6000 GPU. 
To validate the effectiveness of the proposed method in traffic scenarios, we construct a traffic VQA dataset that includes various traffic elements. Specifically, the dataset consists of 41 images of different traffic scenes. Some of these images come from highway scenes in the Figrim dataset~\cite{figrim2015}, while the rest are collected from the internet. All images are captured from the front view of a vehicle, consistent with the driver's perspective. Each image is paired with at least three questions, totaling $N_{total}=172$ questions. Most questions focus on elements such as traffic signs, traffic lights, vehicles, and pedestrians. We use answer accuracy to evaluate whether the model can correctly recognize the information it receives. If the model’s response generally matches the ground-truth answer, it is considered correct; otherwise, it is considered incorrect. The answer accuracy is defined as $\frac {N_{right}} {N_{total}}$, where $N_{right}$ is the number of correct answers.Table~\ref{dataset} provides examples from the dataset.

\begin{table}[ht]
  \centering
  \caption{Example from the traffic VQA dataset.}\label{dataset}
  \begin{tabular}{@{}p{0.9\linewidth}@{}}
    \toprule
    \includegraphics[width=0.65\linewidth]{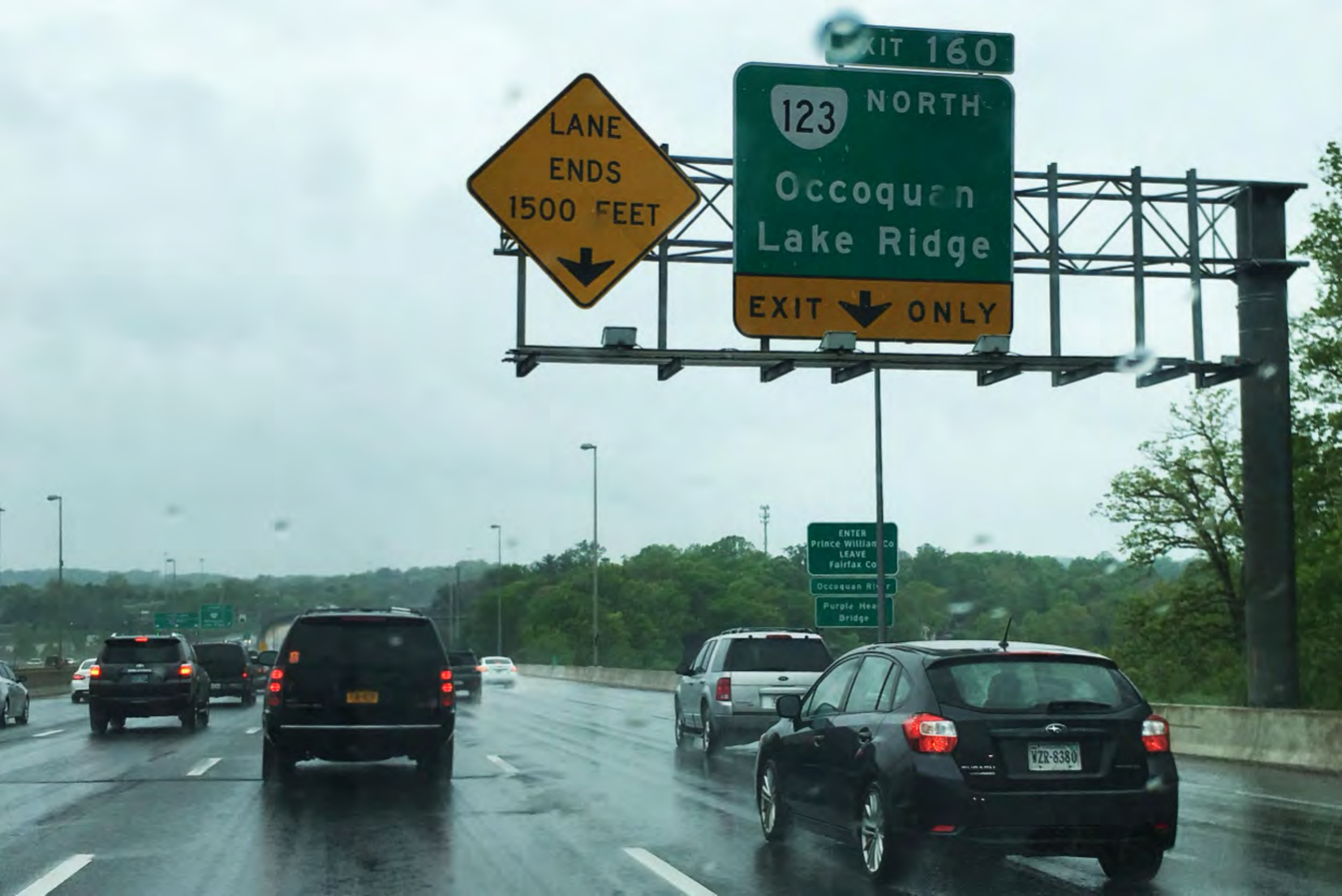} \\
    \midrule
    {\sffamily {Q1}: According to the traffic sign, how do I get to Lake Ridge?} \\
    {\sffamily {Q2}: What is the route number to Occoquan according to the traffic sign?} \\
    {\sffamily {Q3}: What is the exit number ahead according to the traffic sign?} \\
    {\sffamily {Q4}: How is the weather now?} \\
    {\sffamily {Q5}: According to the traffic sign, when does the lane end?} \\
    \midrule
    \includegraphics[width=0.65\linewidth]{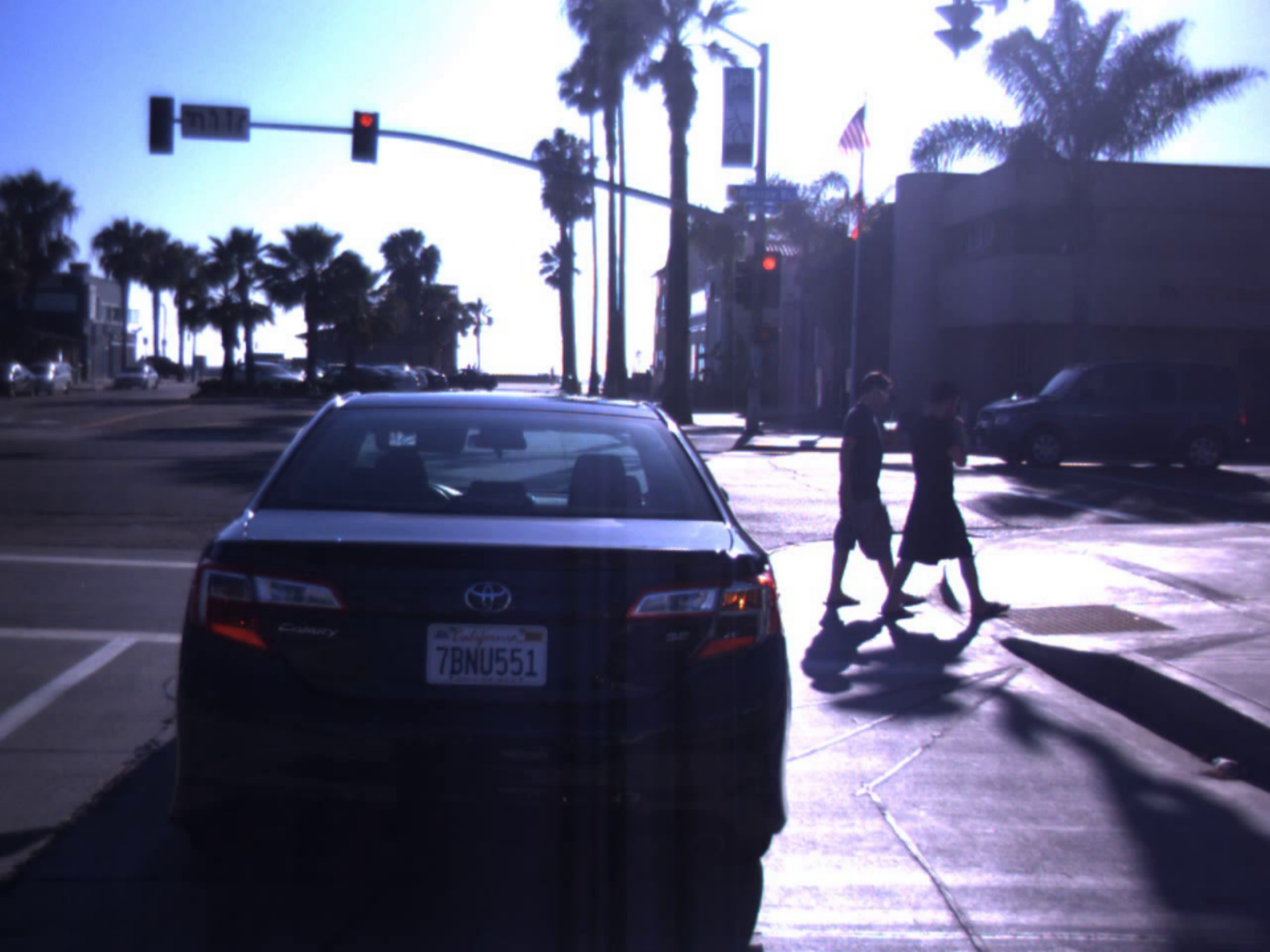} \\
    \midrule
    {\sffamily {Q1}: What should I do given the current traffic light status?} \\
    {\sffamily {Q2}: What color is the traffic light now?} \\
    {\sffamily {Q3}: How many pedestrians are in front?} \\
    {\sffamily {Q4}: What color is the car in front?} \\
    {\sffamily {Q5}: What is the license plate number of the black car in front?} \\
    \bottomrule
  \end{tabular}
\end{table}

To effectively simulate users' objective attention, we employ a saliency prediction model as outlined in reference~\cite{8444709}. This model, which is a cornerstone of the objective attention module, has been pretrained on the MIT1003 dataset~\cite{5459462}. The MIT1003 dataset is recognized as a standard benchmark for evaluating the performance of saliency prediction models, ensuring that the model used in this study is both accurate and efficient in simulating visual saliency across diverse visual scenes. For the SM module, we employ YOLOv8-n~\cite{ultralytics2023yolo} as the object detection model. Its lightweight framework makes it suitable for real-time processing, which is essential for applications that require rapid responses. The model, pre-trained on a dataset of 166 training images and 40 validation images, can accurately identify key categories such as vehicles, traffic signs, traffic lights, and pedestrians.

The channel model utilizes the Fisher-Snedecor \(\mathcal{F}\) model~\cite{Kang2022PersonalizedSI}. Small-scale fading between the vehicle and the server follows the Fisher-Snedecor \(\mathcal{F}\) fading distribution, with small-scale variations and shadowing modeled by the Nakagami-\(m\) and inverse Nakagami-\(m\) distributions, respectively. We set the fading parameter \(m_f = 5\), the Shadowing parameter \(m_s = 4\), and the transmission power \(P = 30\) W by default.

\subsection{Results and Analysis}
\begin{figure}[t]
  \centering
  \includegraphics[width=0.45\textwidth]{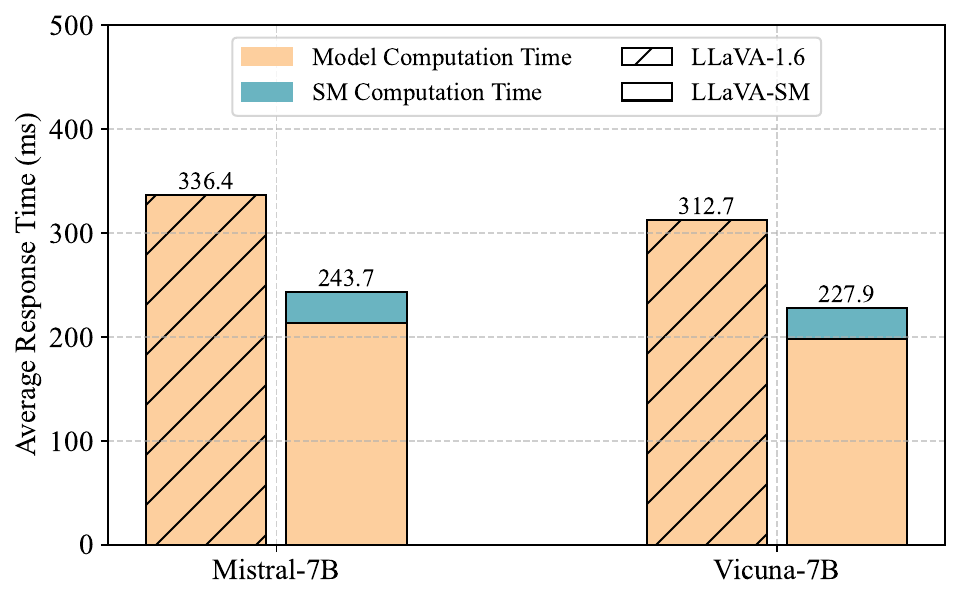}
  \caption{Comparison of average response times for LLaVA-1.6 and LLaVA-SM.}
  \label{latency}
\end{figure}

\textbf{Results of SM model for LLaVA\footnote{Note that LLaVA was not trained on a specialized traffic dataset, which is not the focus of this study, and thus its accuracy is expected to improve significantly after training~\cite{Li2023LLaVAMedTA, Shi2024MathLLaVABM}.}.} To evaluate the performance of the model after integrating the SM module, experiments are conducted using the traffic VQA dataset, and the FLOPs are calculated following the methodology described in~\cite{Zhang2024CLSAI}. The results are shown in Table~\ref{result1}. It can be seen that LLaVA-1.5, which merely scales the original images to the target resolution, struggles to accurately recognize the content of the images, resulting in a model accuracy of only 65.1\%. 
In contrast, LLaVA-1.6, which processes high-resolution images into multiple image slices, not only raises accuracy to 88.4\% (Mistral-7B) and 89.0\% (Vicuna-7B) but also significantly increases the number of image tokens and FLOPs. The SM module introduced in this study reduces the required tokens and FLOPs by using fewer image slices, resulting in only a minimal loss of accuracy. Specifically, Vicuna-7B experiences a decrease of 0.6\%, while Mistral-7B maintains its accuracy. This demonstrates a more efficient balance between performance and computational load. Fig.~\ref{latency} illustrates the average response time per question before and after integrating the SM module. It is evident that the incorporation of the SM module introduces a slight increase in computation time while effectively reducing the overall computational load and the number of image tokens, thereby decreasing the model's response time by approximately 27\%.

\begin{table*}[htbp]
    \centering
    \caption{Efficiency comparison of different models on the traffic VQA dataset.}
    \begin{tabular}{c >{\centering\arraybackslash}m{2.8cm} >{\centering\arraybackslash}m{2.0cm} >{\centering\arraybackslash}m{2.3cm} c}
    \hline
    Model & \# Token & FLOPs (T) & LLM Backbone & Answer Accuracy \\ \hline
    LLaVA-1.5 & 576 & 8.02 & Vicuna-7B & 65.1\% \\ \hline
    \multirow{2}{*}{LLaVA-1.6} & \multirow{2}{*}{2880 ($576 \times 5$)} & \multirow{2}{*}{43.58} & Mistral-7B & 88.4\% \\
     &  &  & Vicuna-7B & 89.0\% \\ \hline
    \multirow{2}{*}{LLaVA-SM} & \multirow{2}{*}{1152 ($576 \times 2$)} & \multirow{2}{*}{17.43} & Mistral-7B & 88.4\% \\
     &  &  & Vicuna-7B & 88.4\% \\ \hline
    \end{tabular}
    \label{result1}
\end{table*}

\begin{figure}[!t]
\centering 
\subfigure[LLM Backbone: Mistral-7B.]{
\label{re11}
\includegraphics[width=0.5\textwidth]{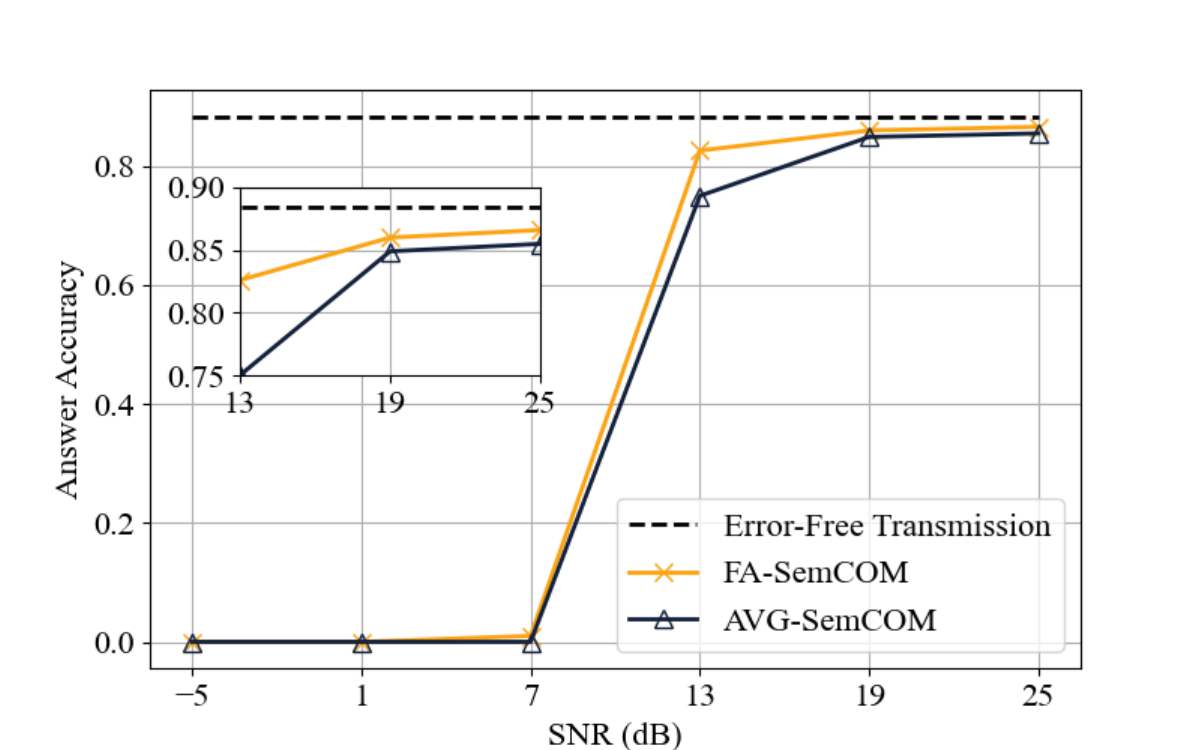}}
\subfigure[LLM Backbone: Vicuna-7B.]{
\label{re12}
\includegraphics[width=0.5\textwidth]{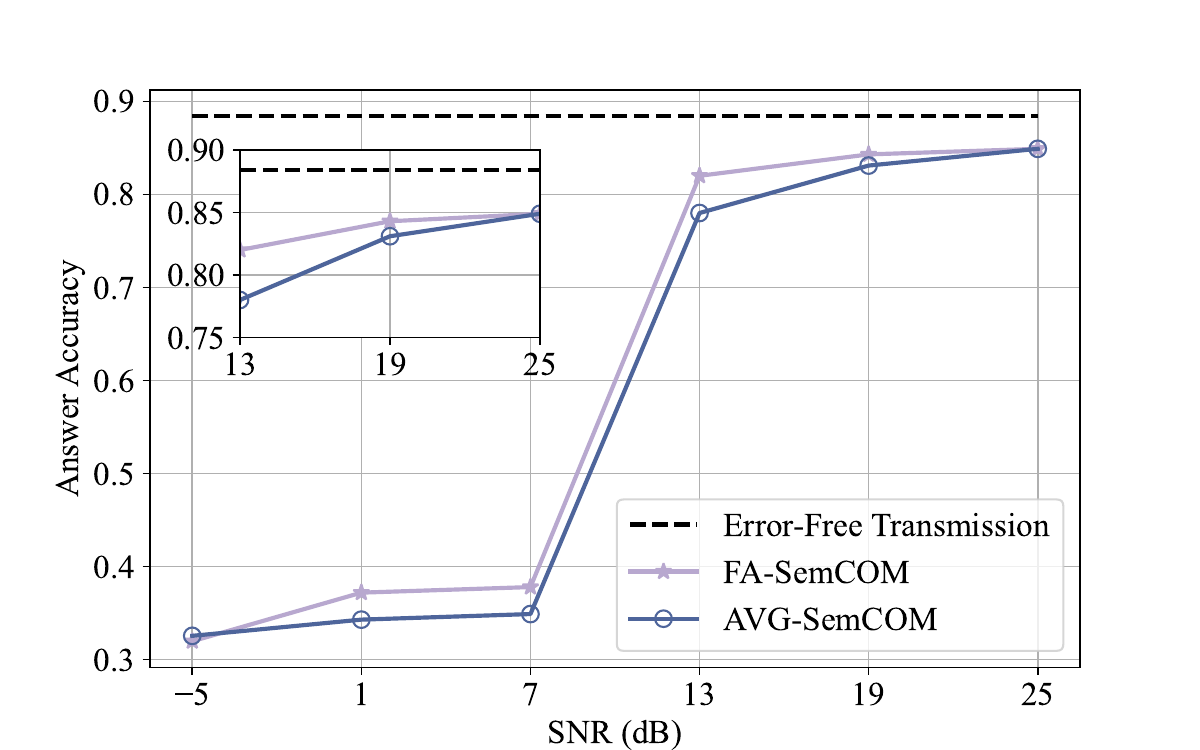}}
\caption{The curves of model answer accuracy for different transmission methods across varying SNR levels.}
\label{re1}
\end{figure}

\textbf{Impact of SNR on Model Performance.}
Fig.~\ref{re1} illustrates the impact of different SNR levels on model performance, where AVG-SemCOM represents the method of average transmission and the weight adjustment variable $\beta$ is set to 1. In this setup, the LLM backbone in Fig.~\ref{re11} is Vicuna-7B, while Fig.~\ref{re12} uses Mistral-7B. The x-axis represents the SNR range from -5 dB to 25 dB, and the y-axis shows the model's response accuracy. From the figures, it is evident that as the SNR decreases, the accuracy of the models also declines, particularly when the SNR drops below 13 dB, where the accuracy of both models sharply decreases. This trend indicates that higher error rates significantly impair the model's performance. However, under low SNR conditions, the Vicuna-7B backbone outperforms the Mistral-7B model. Specifically, Mistral-7B struggles to produce correct outputs when the error rate is high, while Vicuna-7B manages to maintain a certain level of accuracy.

Moreover, the FA-SemCom method shows a noticeable improvement in response accuracy for both language models, especially in poor channel conditions. For instance, at an SNR of 13 dB, the FA-SemCom model's accuracy surpasses that of the AVG-SemCom by 7.6\% and 4\% for the two language models, respectively. It is important to note that $\beta$ has not yet reached its optimal value at this point. This result underscores the importance of considering semantic information in data transmission and highlights the effectiveness of the FA-SemCom model in maintaining accuracy and reliability, particularly in low SNR environments.

\begin{figure}[htbp]
  \centering
  \includegraphics[width=0.45\textwidth]{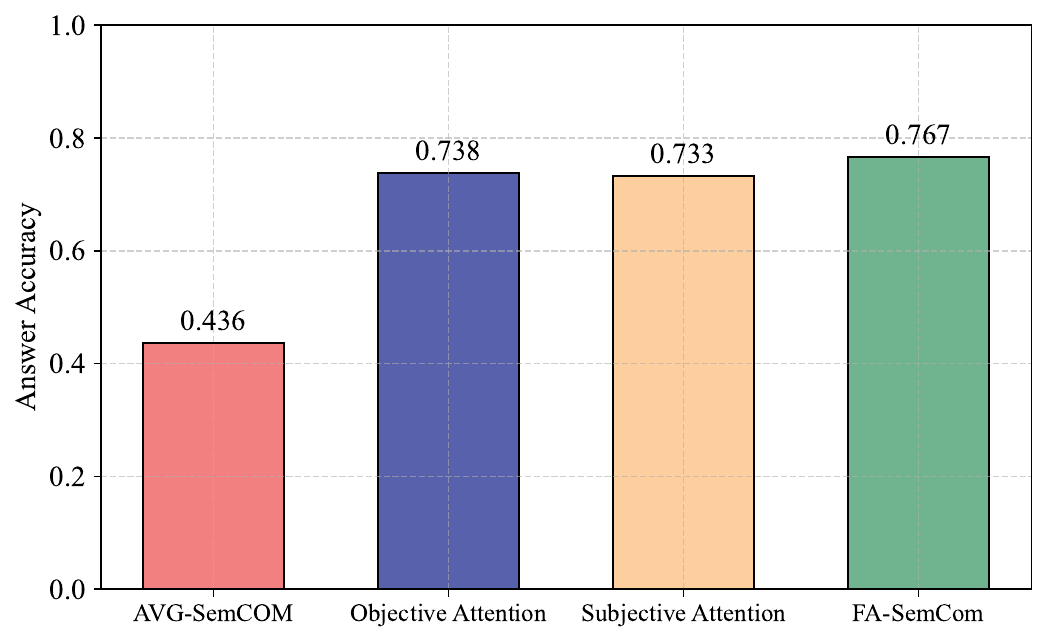}
  \caption{Comparison of different attention mechanisms. The LLM backbone is vicuna-7B with an SNR of 10 dB and $\beta=4$.}
  \label{attention}
\end{figure}

\textbf{The effect of attention mechanisms.}
Fig. \ref{attention} presents the effects of different attention mechanisms on the model’s accuracy during transmission. Without any attention mechanism (i.e., Avg-SemCOM), the accuracy remains relatively low. However, introducing either objective or subjective attention substantially enhances the model’s performance. Moreover, when both types of attention are combined (i.e., FA-SemCOM), the accuracy reaches its highest level, increasing from 0.436 to 0.767.

\begin{figure}[htbp]
  \centering
  \includegraphics[width=0.45\textwidth]{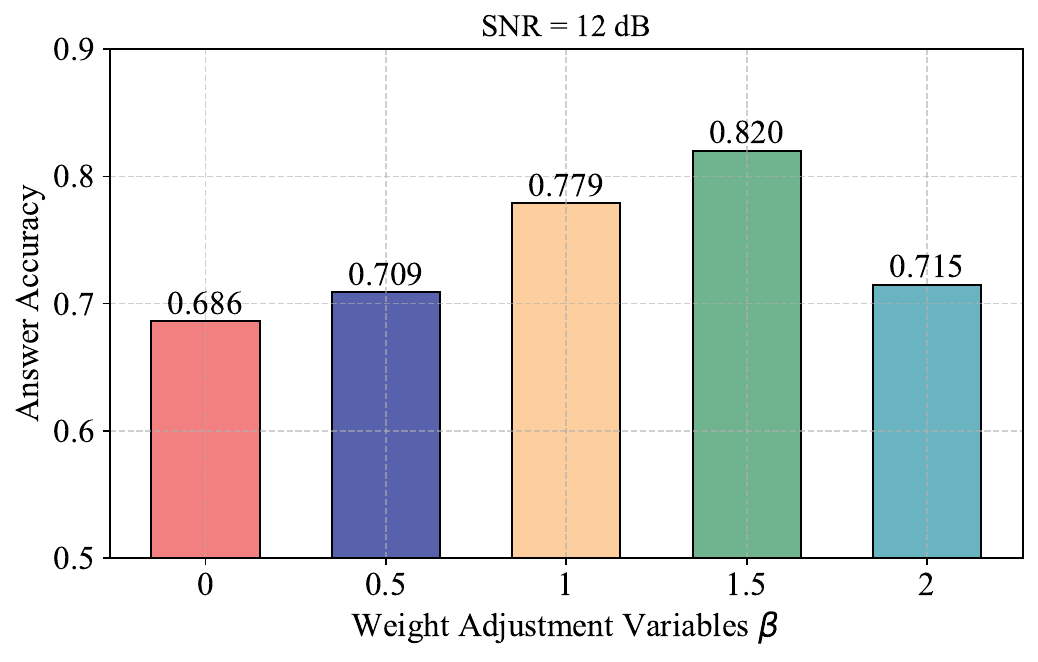}
  \caption{The answer accuracy under different weight adjustment variables $\beta $. The LLM backbone is Mistral-7B with an SNR of 12 dB.}
  \label{re2}
\end{figure}

\textbf{The effect of weight adjustment variables $\beta$.}
We study the impact of weight adjustment variables \(\beta\) on model performance. Fig.~\ref{re2} shows the accuracy of model-generated responses at an SNR of 12 dB. At $\beta=0$, the image feature weights have no differences, indicating an equal distribution of transmission power (i.e., AVG-SemCOM). As $\beta$ increases, the model’s response accuracy improves from 0.686 to 0.802, indicating that assigning greater weight to important features helps mitigate transmission errors. Without such adjustments, the model struggles to correctly interpret images, leading to a relatively low baseline accuracy. However, further increasing $\beta$ beyond a certain point reduces accuracy. Through our experiments, we found that an excessively large $\beta$ value causes the model to produce incorrect or garbled outputs. Hence, in practical scenarios, dynamically adjusting $\beta$ based on channel conditions can enhance both the accuracy and robustness of the model’s responses.

\begin{figure*}[htbp]
    \centering
    \begin{minipage}{0.2\textwidth}
        \centering
        \subfigure[Based Image $X_b$]{
        \includegraphics[width=\textwidth]{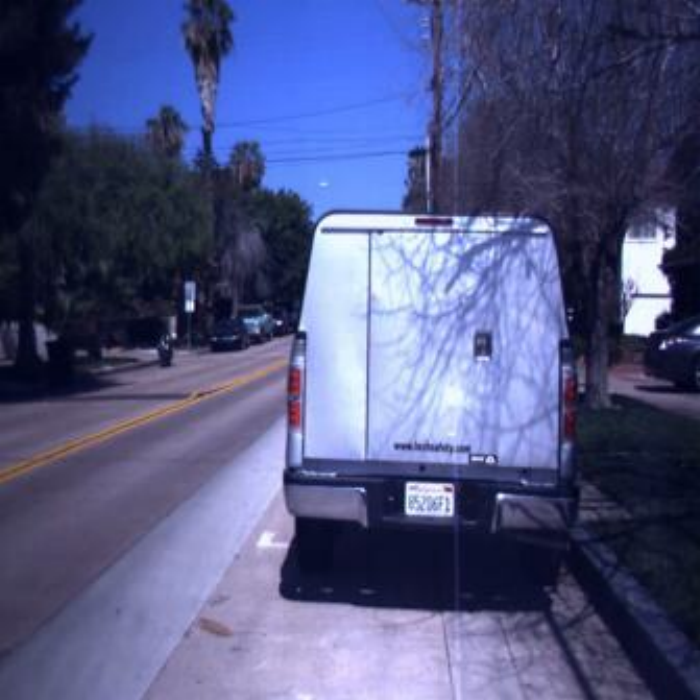}}
    \end{minipage}
    \hspace{5mm}
    \begin{minipage}{0.33\textwidth}
        \centering
        \subfigure[Weight Adjustment Variables $\beta = 0.5$]{
        \includegraphics[width=\textwidth]{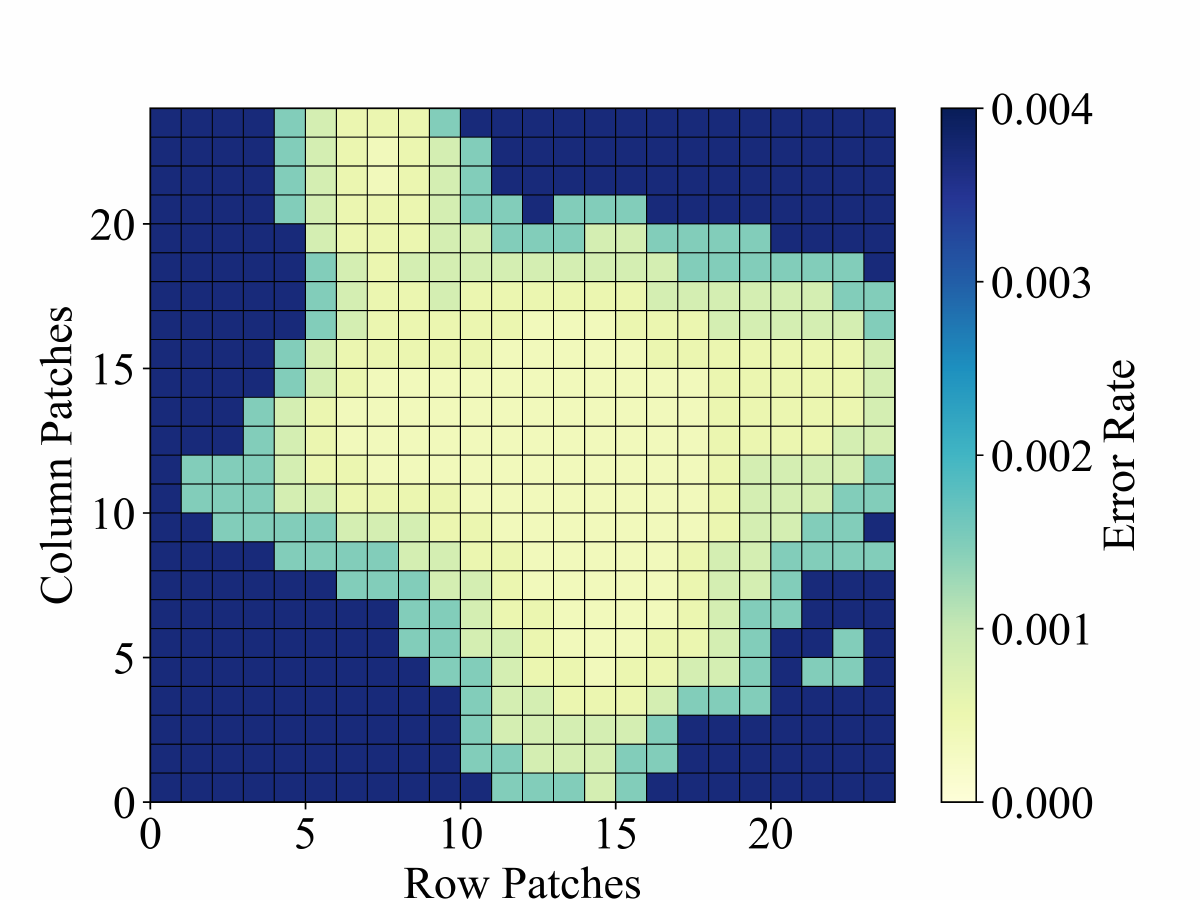}}
    \end{minipage}
    \hspace{3mm}
    \begin{minipage}{0.33\textwidth}
        \centering
        \subfigure[Weight Adjustment Variables $\beta = 1$]{
        \includegraphics[width=\textwidth]{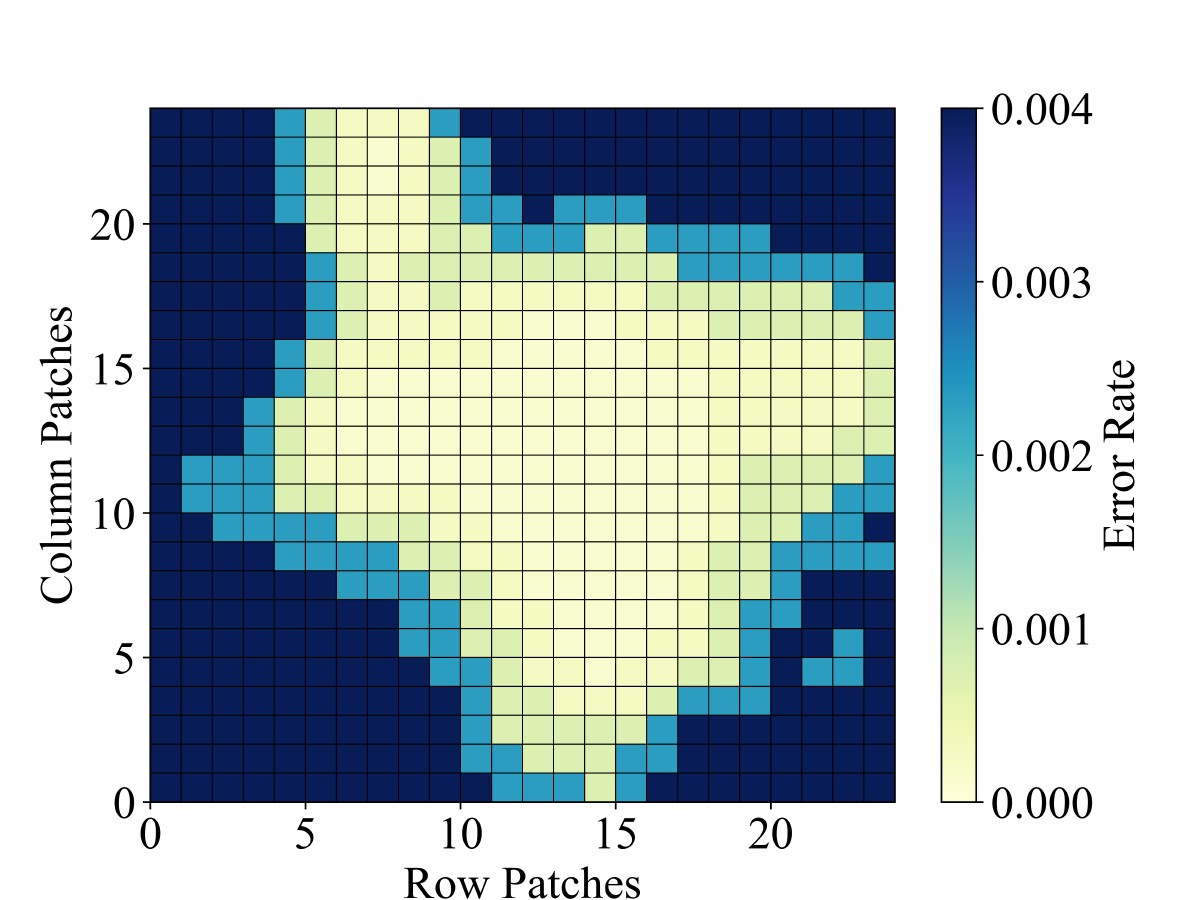}}
    \end{minipage}

    \begin{minipage}{0.2\textwidth}
        \centering
        \subfigure[Image Slice $X_s$]{
        \includegraphics[width=\textwidth]{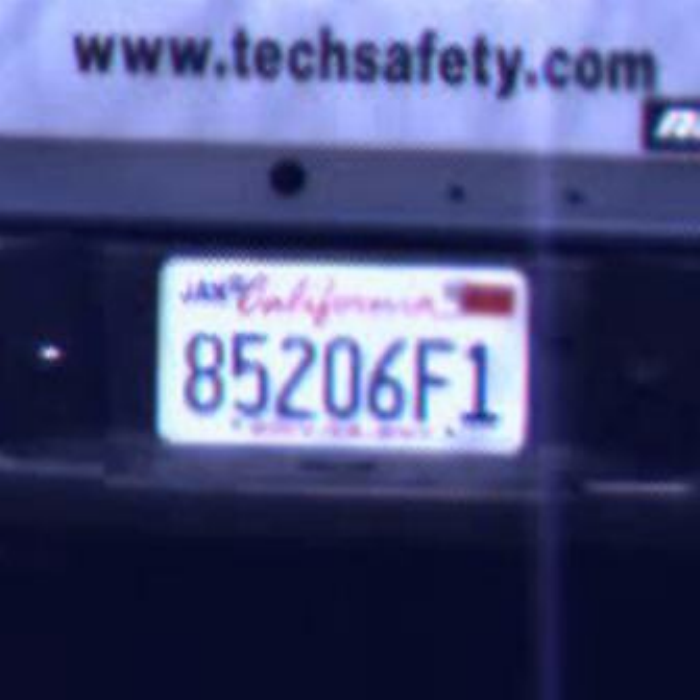}}
    \end{minipage}
    \hspace{5mm}
    \begin{minipage}{0.33\textwidth}
        \centering
        \subfigure[Weight Adjustment Variables $\beta = 0.5$]{
        \includegraphics[width=\textwidth]{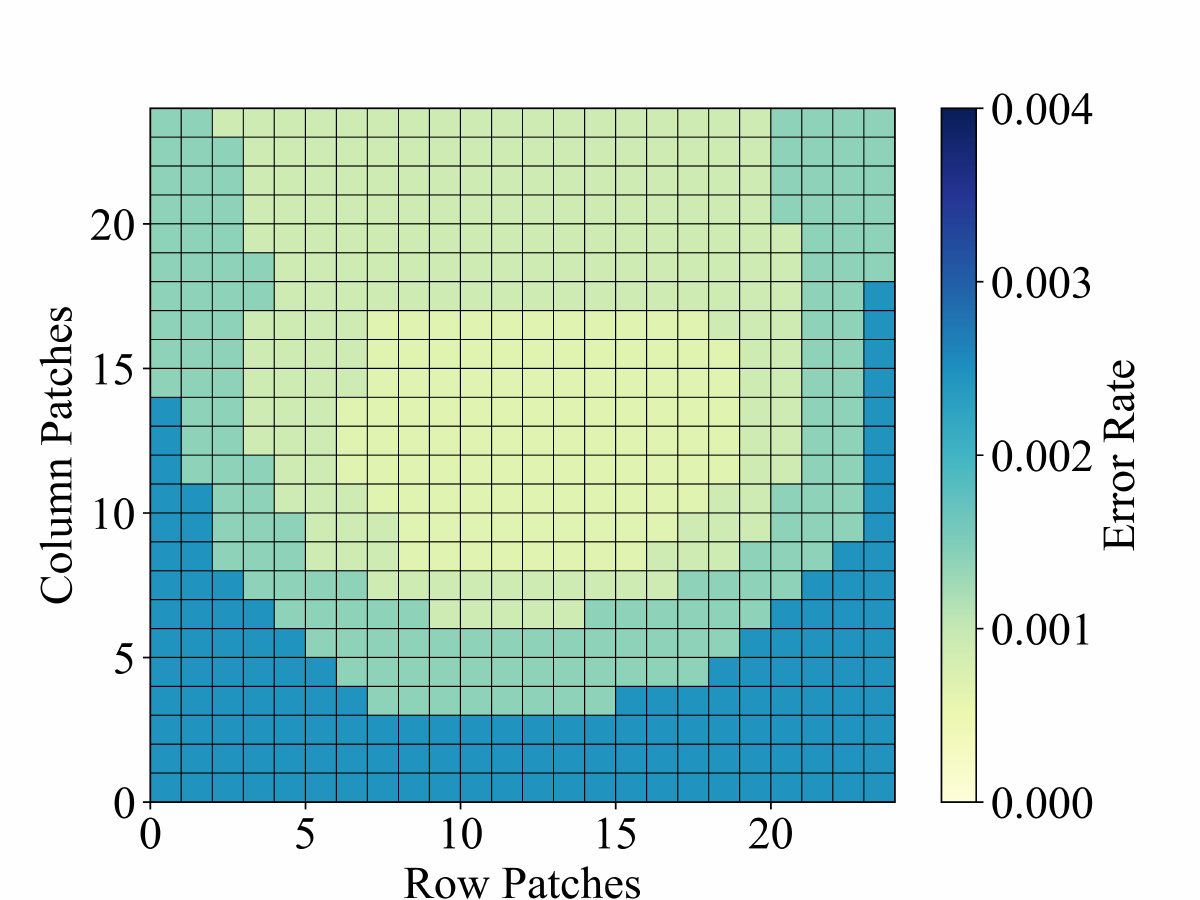}}
    \end{minipage}
    \hspace{3mm}
    \begin{minipage}{0.33\textwidth}
        \centering
        \subfigure[Weight Adjustment Variables $\beta = 1$]{
        \includegraphics[width=\textwidth]{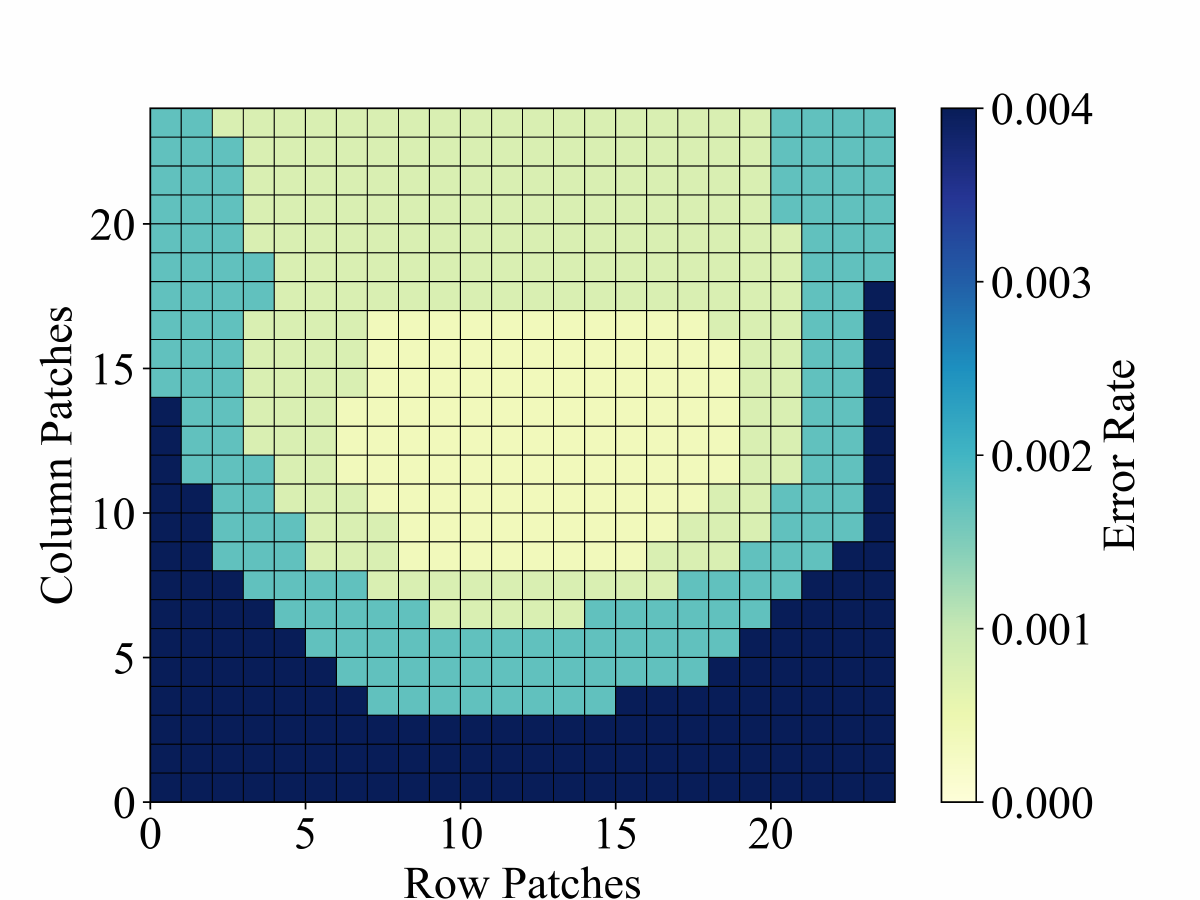}}
    \end{minipage}
    \caption{Visualization of Bit Error Rates (BER) across different image patches with different weight adjustment variables $\beta $ Values, at an SNR of 12 dB.}
    \label{ber2}
\end{figure*}

Additionally, we show the Bit Error Rates (BER) of different image patches at an SNR of 12 dB, with \(\beta\) values set at 0.5 and 1, respectively. As shown in Fig.~\ref{ber2}, when the user's query involves the license plate of a vehicle ahead, the FA-SemCom model allocates more transmission power to the features of the license plate location, ensuring the accurate transmission of this image feature. Moreover, as the $\beta$ value increases, the weight assigned to this area correspondingly rises, enhancing its transmission priority.

Table~\ref{answer} displays the answers generated under different power allocation schemes, where the LLM Backbone is Mistral-7B and SNR = 12 dB. When using the average transmission method, (i.e., AVG-SemCOM), a higher BER leads to the model's inability to accurately recognize the content of the image. However, as the $\beta$ value increases, so does the recognition accuracy. When $\beta=1$, the model can accurately identify this part of the image.

\begin{table}[ht]
  \centering
  \caption{Responses generated by the model using different transmission methods, where the LLM Backbone is Mistral-7B with an SNR of 12 dB.}\label{answer}
  \begin{tabular}{@{}p{0.9\linewidth}@{}}
    \toprule
    {\sffamily {Q}: What is the license plate number of the car in front?} \\
    \midrule
    {\sffamily {AVG-SemCOM:}} \\ 
    {\sffamily \hspace {1.4em}The license plate number of the car in front is \textcolor{red}{5206}.} \\
    {\sffamily {FA-SemCOM ($\beta$  = 0.5):}} \\
    {\sffamily \hspace {1.4em}The license plate number of the car in front is \textcolor{red}{855206}.} \\
    {\sffamily {FA-SemCOM ($\beta$ = 1):}} \\
    {\sffamily \hspace {1.4em}The license plate number of the car in front is \textcolor{green}{85206F1}.} \\
    \bottomrule
  \end{tabular}
\end{table}

\section{Discussion}\label{S7}
The applicability of our proposed framework extends beyond vehicle communication scenarios. Benefiting from the shared architectural foundations among different LMMs~\cite{Yin2023ASO}, the FA-SemCom method can flexibly adapt to various domains by retraining the semantic module and saliency prediction model on domain-specific datasets~\cite{Li2023LLaVAMedTA}. For instance, in agricultural crop disease detection, selectively identifying and prioritizing the transmission of lesion-critical areas could notably enhance model accuracy in bandwidth-constrained environments. Nevertheless, expanding the framework to other application domains may encounter the following considerations:
\begin{itemize}
    \item \textit{Resource Constraints:} Edge and remote devices typically have stringent computational and energy limitations, necessitating lightweight models or intelligent offloading strategies to optimize efficiency. Dynamically adjusting the depth of image encoding layers according to available computational resources can help balance performance and energy consumption.
    \item \textit{Precision Requirements:} Complex application domains (e.g., agriculture, aviation) often involve intricate backgrounds or small-sized targets, making precise detection challenging. Advanced detection models or domain-specific fine-tuning may be required to effectively capture subtle, critical features.
    \item \textit{Real-time Constraints:} While our current framework meets most real-time demands, latency becomes critical in scenarios with stringent time requirements, especially when deploying large-scale LMMs. Future research should investigate more efficient acceleration techniques—such as computation caching, token compression, and model quantization~\cite{10.1162/tacl_a_00704}—to further reduce inference latency and enhance real-time responsiveness.
\end{itemize}

\section{Conclusion}\label{S8}

In this paper, we have explored task-oriented semantic communication for LMM-based vehicle artificial intelligence systems. We deploy LLaVA's image encoder on the vehicle and position the computationally intensive LLM on a server, transmitting encoded image features as semantic information. To reduce computational load and shorten response time, we propose a SM-based image slicing method, effectively decreasing the number of image slices that require processing. Additionally, by integrating objective and subjective user attention, we quantify semantic importance to precisely prioritize feature transmission, complemented by a fused attention-based power allocation strategy.
Experimental results on a traffic VQA dataset demonstrate significant accuracy improvements, particularly in low SNR scenarios. Our proposed approach surpasses the average transmission method by 13.4\% at 12 dB and 33.1\% at 10 dB.

\bibliographystyle{IEEEtran}
\bibliography{Ref}

\begin{IEEEbiographynophoto}{Baoxia Du} (Student Member, IEEE)
     is a PhD student in College of Science and Engineering, Kanazawa University, Japan. He received the B.S. degree from Shandong University of Science and Technology, Qingdao, China, in 2020, and the M.S. degree from the School of Information and Control Engineering, Jilin Institute of Chemical Technology, Jilin City, China in 2024. His current research interests include digital twin, deep learning, computer vision, and semantic communication.
    \vspace{-20pt}
\end{IEEEbiographynophoto}

\begin{IEEEbiographynophoto}{Hongyang Du} (Member, IEEE) is an assistant professor at the Department of Electrical and Electronic Engineering, The University of Hong Kong. He received the BEng degree from the Beijing Jiaotong University, Beijing, and the PhD degree from the Nanyang Technological University, Singapore. He serves as the Editor-in-Chief Assistant (2022-2024) and Editor (2025-Present) of IEEE Communications Surveys \& Tutorials, the Editor of IEEE Transactions on Communications, IEEE Transactions on Vehicular Technology, IEEE Open Journal of the Communications Society, and the Guest Editor for IEEE Vehicular Technology Magazine. He is the recipient of the IEEE ComSoc Young Professional Award for Best Early Career Researcher in 2024, IEEE Daniel E. Noble Fellowship Award from the IEEE Vehicular Technology Society in 2022, and the IEEE Signal Processing Society Scholarship from the IEEE Signal Processing Society in 2023. He was recognized as an exemplary reviewer of the IEEE Transactions on Communications and IEEE Communications Letters. His research interests include edge intelligence, generative AI, and network management.
\vspace{-20pt}
\end{IEEEbiographynophoto}

\begin{IEEEbiographynophoto}{Dusit Niyato} (Fellow, IEEE) received B.Eng. from King Mongkut's Institute of Technology Ladkrabang (KMITL), Thailand in 1999 and the Ph.D. in Electrical and Computer Engineering from the University of Manitoba, Canada in 2008. He is currently a professor in the College of Computing and Data Science, at Nanyang Technological University, Singapore. His research interests are in the areas of the Internet of Things (IoT), machine learning, and incentive mechanism design.
\vspace{-20pt}
\end{IEEEbiographynophoto}

\begin{IEEEbiographynophoto}{Ruidong Li (Corresponding author)} (Senior Member, IEEE) received the Ph.D. degree from University of Tsukuba, Japan in 2008. He was with NICT, Japan from 2008 to 2021 and joined Kanazawa University, Japan as an associate professor since 2021. He is the recipient of the best paper awards for IEEE ICC 2022 and IWCMC 2022, IEEE Network 2022 Editor Appreciation Award, and IEEE MMTC Outstanding Leadership Award 2023. He serves as the chair of IEEE ComSoc Internet Technical Committee (ITC) and founding chairs for IEEE SIG on Big Data Intelligent Networking, Intelligent Internet Edge, and Metaverse.

\end{IEEEbiographynophoto}

\end{document}